\documentclass[journal]{IEEEtran}

\usepackage[english]{babel}
\usepackage{amsmath}
\usepackage{graphicx}
\usepackage[colorlinks=true, allcolors=black]{hyperref}
\usepackage{multirow, makecell}
\usepackage{authblk}
\usepackage{ulem}
 
\usepackage{tabularx}

\newcolumntype{Y}{>{\centering\arraybackslash}X}
\newcolumntype{P}[1]{>{\centering\arraybackslash}p{#1}}
\newcolumntype{k}{>{\hsize=.2\hsize}Y}

\usepackage{cite}

\begin{document}
\title{Rain regime segmentation of Sentinel-1 observation learning from NEXRAD collocations with Convolution Neural Networks}

\author{Aurélien~Colin$^{1,2}$,
        Pierre~Tandeo$^{1}$,
        Charles~Peureux$^{2}$,
        Romain~Husson$^{2}$,
        Nicolas~Longépé$^{3}$,
        Ronan~Fablet$^{1}$,
\thanks{$^1$ IMT Atlantique, Lab-STICC, UMR CNRS 6285, F-29238, France.}
\thanks{$^2$ Collecte Localisation Satellites, Brest, France.}
\thanks{$^3$ $\Phi$-lab Explore Office, ESRIN, European Space Agency (ESA), Frascati, Italy}}

\markboth{Journal of \LaTeX\ Class Files,~Vol.~13, No.~9, September~2014}%
{Shell \MakeLowercase{\textit{et al.}}: Bare Demo of IEEEtran.cls for Journals}

\maketitle

\begin{abstract}
Remote sensing of rainfall events is critical for both operational and scientific needs, including for example weather forecasting, extreme flood mitigation, water cycle monitoring, etc.
Ground-based weather radars, such as NOAA's Next-Generation Radar (NEXRAD), provide reflectivity and precipitation estimates of rainfall events. However, their observation range is limited to a few hundred kilometers, prompting the exploration of other remote sensing methods, particularly over the open ocean, that represents large areas not covered by land-based radars.
Here we propose a deep learning approach to \textcolor{black}{deliver a three-class segmentation of SAR observations  in terms of rainfall regimes}. SAR \textcolor{black}{satellites deliver very high resolution observations with a} global coverage. \textcolor{black}{This seems particularly appealing to inform fine-scale rain-related patterns, such as those associated with convective cells with characteristic scales of a few kilometers.}
We demonstrate that a convolutional neural network trained on a collocated Sentinel-1/NEXRAD dataset clearly outperforms state-of-the-art filtering schemes such as the Koch's filters.
Our results indicate high performance in segmenting precipitation regimes, delineated by thresholds at 24.7, 31.5, and 38.8 dBZ. Compared to current methods that rely on Koch's filters to draw binary rainfall maps, these multi-threshold learning-based models can provide rainfall estimation. They may be of interest 
\textcolor{black}{in improving high-resolution SAR-derived wind fields, which are degraded by rainfall, and provide an additional tool for the study of rain cells.}
\end{abstract}


\IEEEpeerreviewmaketitle

\section{Introduction}

\IEEEPARstart{P}{recipitation} monitoring and forecasting are major operational and scientific challenges. 
\textcolor{black}{The monitoring of rainfall and convective systems can greatly benefit from high-resolution maps derived through remote sensing techniques \cite{10.1029/2019ms001618}. This is particularly relevant within the context of climate change, as numerous coastal regions are projected to encounter increased precipitation and a higher frequency of extreme rainfall events  \cite{douville2021}.
Real-time satellite data shows promise for flash flood nowcasting \cite{10.3390/rs5115702, 10.1109/tgrs.2007.903685}.
However, the analysis of rain cells necessitates a wide area of observation with a high degree of detail, particularly in the case of convective cells which usually measure a few dozen kilometers in diameter \cite{10.1016/j.rse.2015.02.006}, though they can be smaller in the early stages of development \cite{ayet2021uncovering}.}

Ground-based weather radars provide high-resolution rainfall estimates that are limited \textcolor{black}{in} coastal areas due to their range that spans only a few hundred kilometers. 
\textcolor{black}{Among other parameters, }
such radars measure the reflectivity of the air column at different inclinations. When the beam encounters precipitation, part of the emitted signal is reflected back to the sensor with an intensity that depends on \textcolor{black}{ the size of the water droplets, their density, and the distance from the radar}. Previous studies have derived empirical relationships between weather radar reflectivities, the categories of the hydrometeors and the rainfall rates \cite{10.1175/1520-0469, 10.1175/jhm-d-19-0194.1}. 
The Next Generation Weather Radar (NEXRAD) systems, used in this study, have a resolution of 1 km in range and 1° in azimuth.

Farther from the coast, satellite-derived rain products are available,
 though at a lower spatial resolution than ground-based radars. Sensors deployed on low-Earth orbit satellites provide coverage of the entire globe at an extended temporal resolution. For example, brightness temperatures measured by microwave radiometers \cite{10.1016/b0-12-227090-8}, such as SSMI/S, can provide rain rate estimates. SSMI/S' along-track and cross-track resolution is respectively 14 and 13 km/px.
Satellite-based radars such as GPM-DPR \cite{10.2151/jmsj.2021-011} are also available, with an higher resolution of 5 km/px. Satellite observations can be merged into multi-satellite products such as GPM IMERG \cite{10.1007/978-3-030-24568-9_19} or CMORPH \cite{10.1175/1525-7541(2004)005<0487:CAMTPG>2.0.CO;2}. GPM IMERG uses low-orbit infrared (IR) observations with a spatial resolution of 0.1° and a temporal resolution of 30 minutes. CMORPH also uses IR observations but from geostationary satellites and provides a rainfall product at a spatial resolution of 8 km and a temporal resolution of 30 minutes.

Space-based synthetic aperture radar (SAR) observations measure the backscattered radar signal at high resolution, typically 10 to 25 m for Sentinel-1. \textcolor{black}{They} provide \textcolor{black}{sea surface} images \textcolor{black}{which reveal} a wide variety of meteorological and atmospheric phenomena \cite{jackson2004synthetic}. Among these, rain signatures appear as light and/or dark spots (Fig.\ref{fig:annexe2}). Studied for a long time now, these signatures can be a combination of different contributions from the roughness of the sea surface (increased or decreased surface scattering) or from the atmosphere (volume scattering or attenuation by hydrometeors).
Their impact on radar backscatter varies as a function of many parameters such as incidence angle, wind conditions, signal polarization and frequency, or precipitation rate \cite{10.1109/IGARSS.1996.516666, 10.1117/12.373044, 10.1109/36.921411, 10.1029/2000JC000263}. \textcolor{black}{These rain signatures often hinder other SAR-based information such as the wind field \cite{Portabella2001, 10.1029/JC095iC10p18353}.}
\textcolor{black}{The C-band SAR instrument on} Sentinel-1 satellites can operate in different acquisition modes. One such mode is the Interferometric Wide Swath (IW) mode. It covers several hundred kilometers in range and azimuth directions and extends over incidence angles between 29° and 46°. These products are generally used to study coastal areas. \textcolor{black}{The revisit period of a single Sentinel-1 satellite is between 2 and 6 days depending on the latitude \cite{10.1109/igarss.2008.4778801}.}

\textcolor{black}{
Methods have been developed to detect rainfall from SAR instruments in the Ku-band \cite{10.1109/tgrs.2008.2001032} or X-band \cite{10.1109/tgrs.2019.2953143}}. Although different studies have addressed the estimation of rainfall in the C-band \cite{10.1109/tgrs.2014.2367654} or the segmentation of rain cells \cite{10.1002/gdj3.73,Colin2022}, the calibration of SAR-derived rainfall products remains a challenge \cite{10.1016/j.rse.2016.10.015}. The lack of SAR dataset with groundtruthed reference rainfall data has certainly been a critical limitation. However, Earth observation systems, such as the Sentinel-1 satellites, now \textcolor{black}{deliver} large-scale datasets of SAR observations combined with rainfall data provided by weather radars, specifically NOAA's NEXRAD sensors \cite{rs13163155}. Therefore, the preparation and analysis of well collocated NEXRAD and Sentinel-1 measurements, both at high resolution, provides a unique opportunity to better characterize and detect rain signatures in SAR acquisitions. \textcolor{black}{This methodology is not restricted to Sentinel-1 and can be utilized for other C-band SAR missions like Gao Fen 3 \cite{10.3390/rs10121929}, the RADARSAT Constellation Mission \cite{10.1080/07038992.2015.1104633}, or RISAT \cite{10.1109/ursigass.2014.6929612}.}
 
In this study, we \textcolor{black}{explore} how deep learning \textcolor{black}{approaches} can leverage such a large-scale SAR-NEXRAD dataset \textcolor{black}{to deliver} SAR-derived rainfall estimation. We focus specifically on vertical-vertical (VV) polarization, available for all \textcolor{black}{non-polar} Sentinel-1 products. 
\textcolor{black}{Our underlying assumption is that the relationship between the sea surface roughness and the reflectivity within the air column is strong enough to infer reflectivity bins from the former.} We show that a U-Net architecture far outperforms the filtering-based schemes previously suggested in \cite{rs13163155}. 


\section{Dataset}

The Sentinel-1 mission consists of two satellites, Sentinel-1A and Sentinel-1B, whose synthetic aperture radars (SAR) regularly acquire data at 5.4 GHz (C-band). In this study, we used the IW acquisition mode. IW Ground Range Detected High Resolution (GRDH) products are obtained with a pixel spacing of 10 x 10 meters and a spatial resolution of approximately 20 x 22 meters. These products extend over a few hundred kilometers in range (250 km) and in azimuth.  

\subsection{\textcolor{black}{Rainfall information}}

\textcolor{black}{In preliminary studies, c}ollocation with the satellite-based radar GPM/DPR, conducted over a global Sentinel-1 dataset found only 2,304 partial collocations out of 182,153 IW. 'Partial collocations' is meant to indicate that at least 20x20 km of a swath is observed by \textcolor{black}{GPM} 20 minutes before or after the SAR observation, disregarding the occurrence of rain events. \textcolor{black}{The study also considered collocation with two global products: GPM-IMERG \cite{10.1007/978-3-030-24568-9_19} and CMORPH \cite{10.1175/1525-7541(2004)005<0487:CAMTPG>2.0.CO;2}, both integrated from multiple satellites. Nevertheless, due to the brief lifespan and rapid evolution of convective cells, it is imperative to minimize the time gap between SAR observations and rainfall data acquisition. Additionally, given that rain cells can span only a few kilometers, a high spatial resolution for rainfall data is indispensable.}

The difficulty to obtain collocations, the lower resolution (\textcolor{black}{respectively 5 km, 0.1° and 8 km per pixel for GPM/DPR GPM IMERG and CMORPH}) and greater misalignment issues led \textcolor{black}{us} to \textcolor{black}{rather} focus on \textcolor{black}{weather radar data} as the source of rainfall information. \textcolor{black}{We collected w}eather radar reflectivity \textcolor{black}{data} from NEXRAD, a network of 160 Doppler weather radars operating between 2.7 and 3 GHz. We used the basic reflectivity with a resolution of 1 km in range and 1° in azimuth. Basic reflectivity is provided with a time resolution of \textcolor{black}{6} minutes\textcolor{black}{, meaning that the maximum time difference between Sentinel-1 and NEXRAD observations is 3 minutes}.

\subsection{Sea surface wind fields}

In the absence of rainfall, wind speed is the primary parameter governing variations in sea surface roughness\textcolor{black}{. H}eavy precipitation \textcolor{black}{however} strongly impacts the latter \cite{10.1016/j.rse.2019.111457}. To ensure that the training, validation and test subset follow similar distribution\textcolor{black}{s}, we choose to balance them with respect to both the rainfall and the wind speed. \textcolor{black}{This requires to complement our dataset} with the most reliable wind speed information.

Two sources of wind speed were considered: European Centre for Medium-Range Weather Forecast (ECMWF) model estimates and wind inversions from SAR observations.

\begin{itemize}
    \item The atmospheric model estimates were obtained from sea surface wind at a height of 10 m through the analysis of the ECMWF's global model: the Integrated Forecasting System (IFS). Forecasts are provided either 3-hourly or hourly and at resolutions of 0.125° or 0.1°, respectively before or after August 2019. 
    \item The SAR-derived sea surface wind field relies on CMOD5.N (a C-band geophysical model function) and auxiliary ECMWF wind data. These data are used as priors in a Bayesian inversion scheme \cite{owi}.
\end{itemize}

Significant differences between the SAR-derived and model-derived wind fields can occur if the model is not well phased with respect to the actual situation, if the difference between the analysis and observation times is too large or because of the inherent difference between the resolution of the two wind fields. Other significant differences can occur when the sea surface roughness is impacted by non-wind processes. As highlighted in \cite{10.1016/j.rse.2016.10.015}, four \textcolor{black}{main} physical processes contribute to the radar signature of rainfall events: 1) scattering of the radar signal from the sea surface, the roughness of which is altered by both ring wave generation and wave damping due to turbulence caused by raindrops hitting the sea surface, 2) increased sea surface roughness due to downdraft winds often associated with rain cells, 3) scattering from splash products, i.e. craters, stalks, crowns and rain drops bouncing upwards, and 4) scattering and attenuation of the radar pulse by raindrops (hydrometeors) in the atmosphere (volume scattering and attenuation) which can become non-negligible at very high rain rates. \textcolor{black}{The d}irect interpretation of the sea surface roughness as being a result of sea surface winds would lead to significant errors. An incorrect a priori wind direction model can also lead to incorrect estimates of SAR-derived wind speed and direction. This is generally the case for fast moving phenomena or local structures not seen by modeled winds.

\begin{figure*}[!ht]
    \centering
    \resizebox{\linewidth}{!}{%
    \begin{tabularx}{\linewidth}{kYY}

    & 2018-05-17 23:05:21 & 2019-08-27 22:19:26 \\
    
    \rotatebox[origin=t]{90}{ \parbox{1.8cm}{\centering SAR\\observation }}
    &\includegraphics[width=1\linewidth]{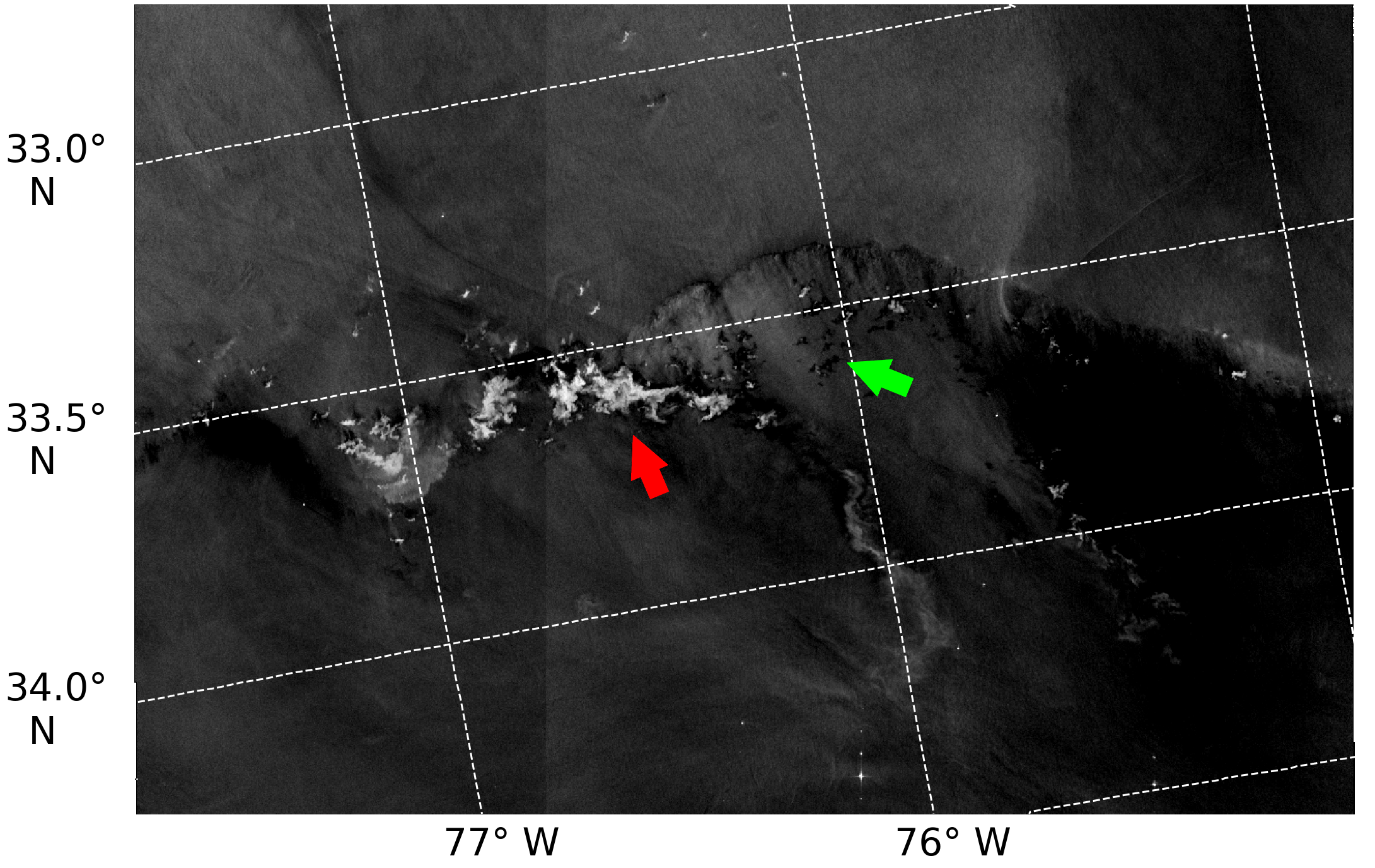}
    & \includegraphics[width=1\linewidth]{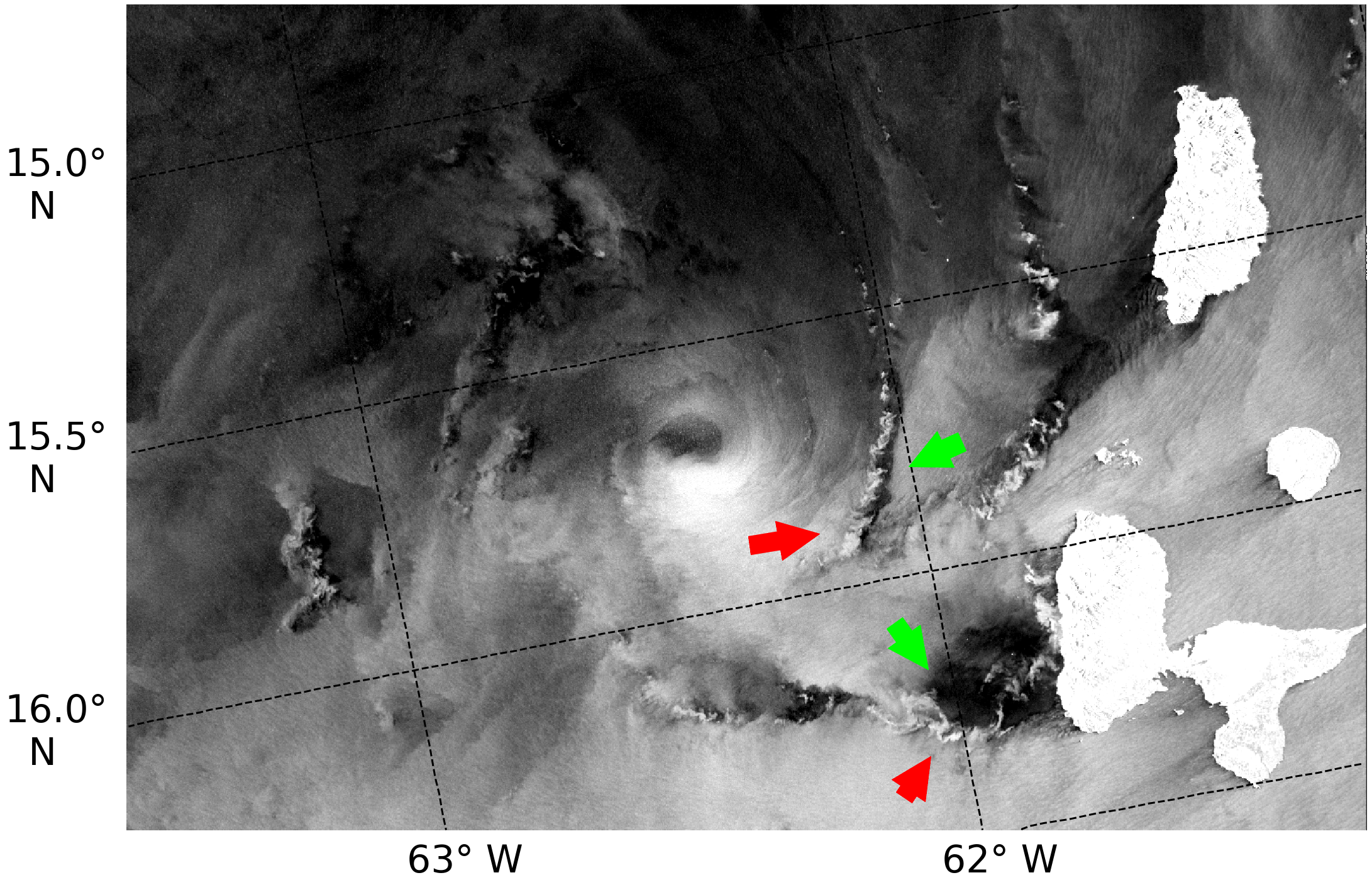}\\
    
    & \multicolumn{2}{c}{\includegraphics[width=0.45\linewidth]{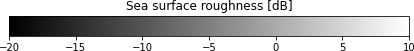}}\\
    
    \rotatebox[origin=t]{90}{ \parbox{1.8cm}{\centering SAR-based\\wind speed}}
    &\includegraphics[width=1\linewidth]{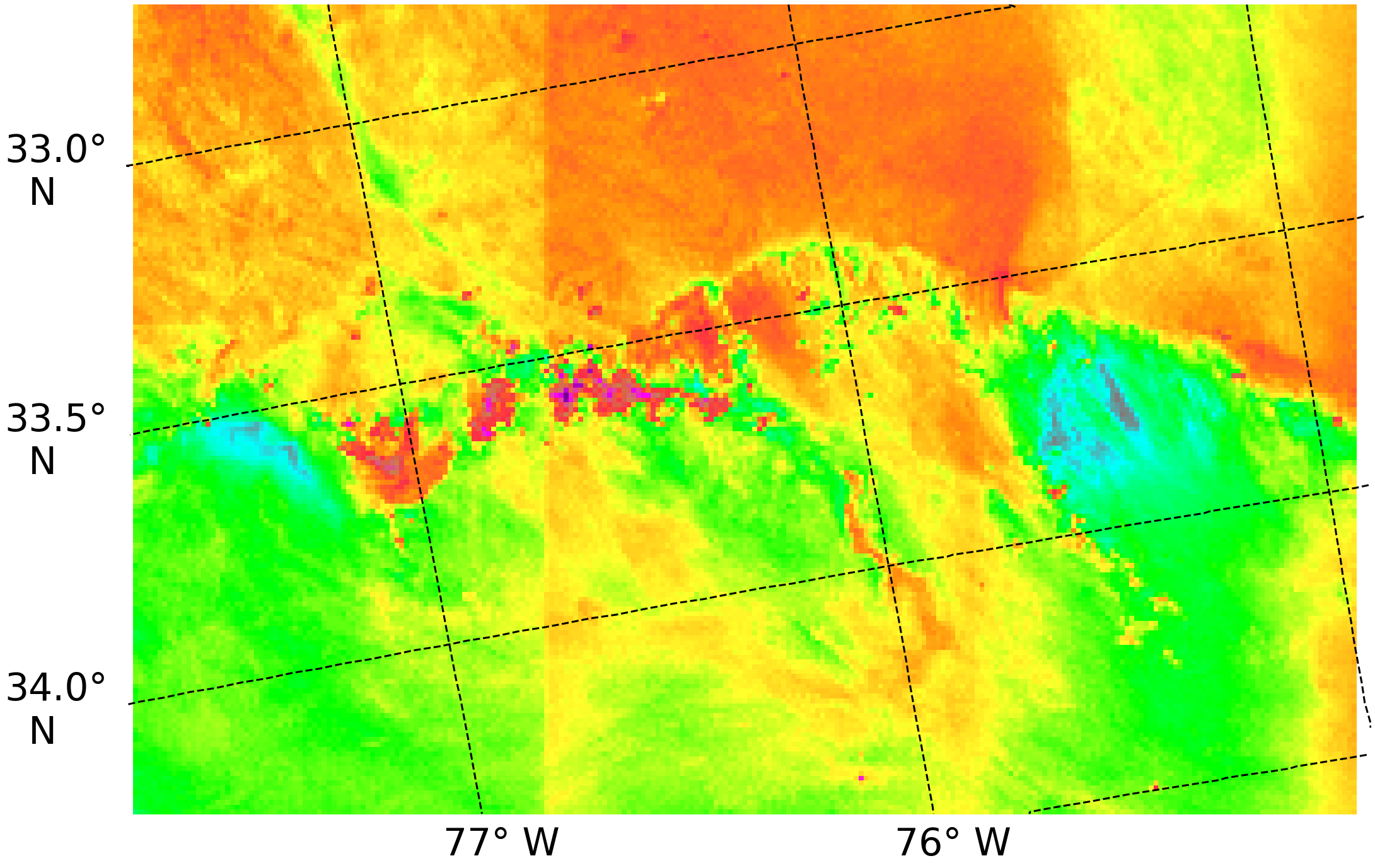}
    & \includegraphics[width=1\linewidth]{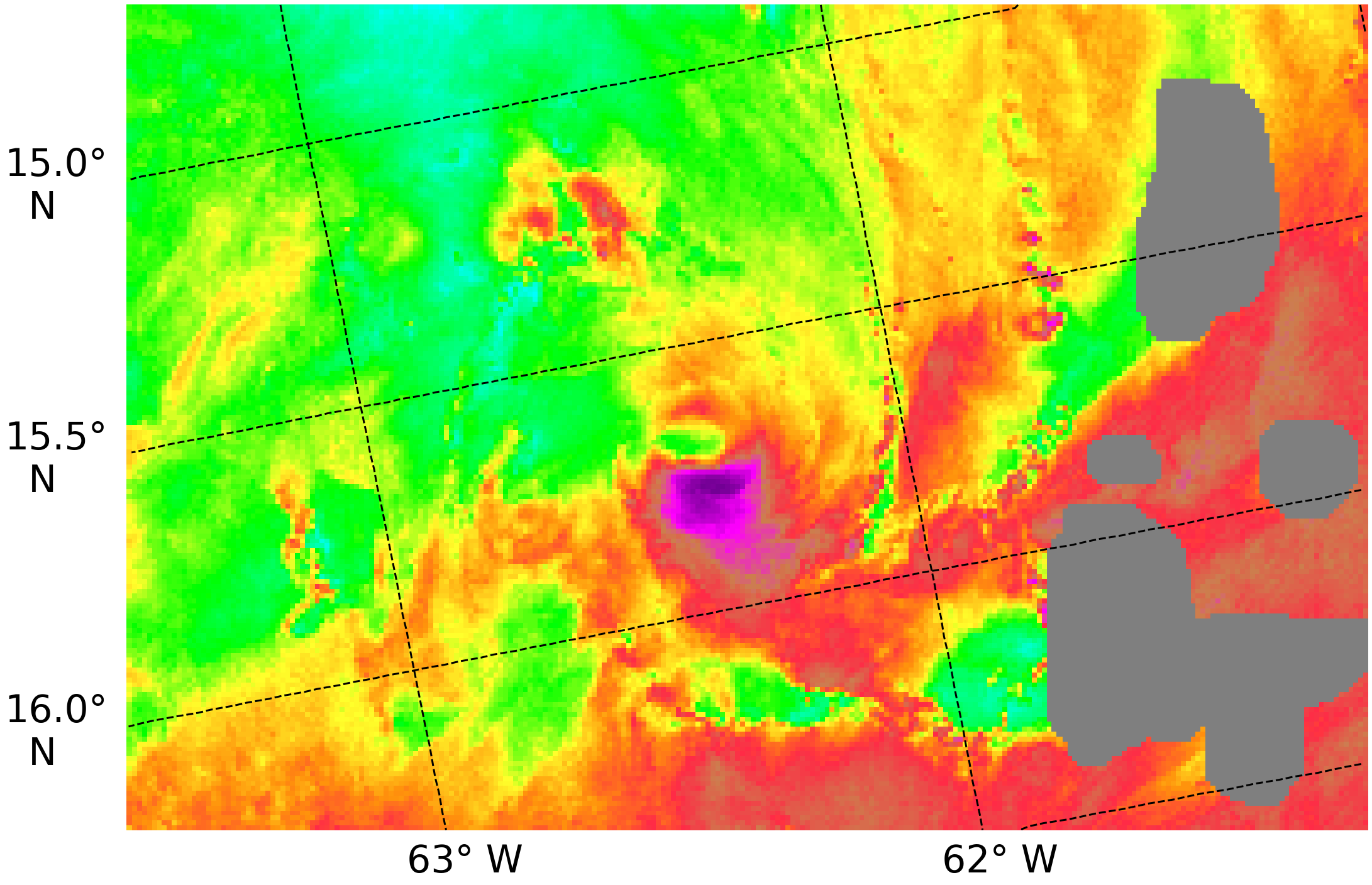}\\
    
    \rotatebox[origin=t]{90}{ \parbox{1.8cm}{\centering ECMWF\\wind speed}}
    &\includegraphics[width=1\linewidth]{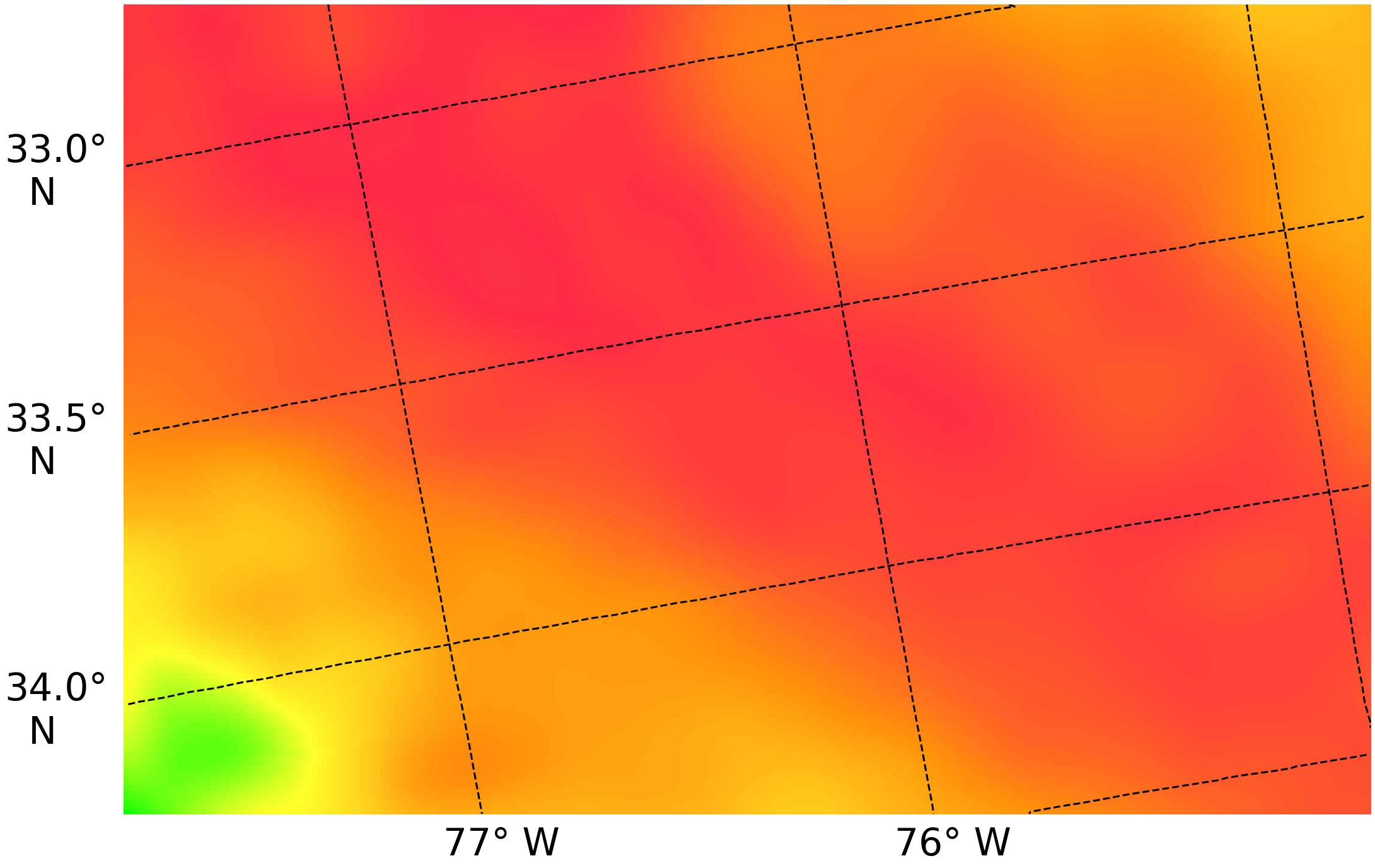}
    & \includegraphics[width=1\linewidth]{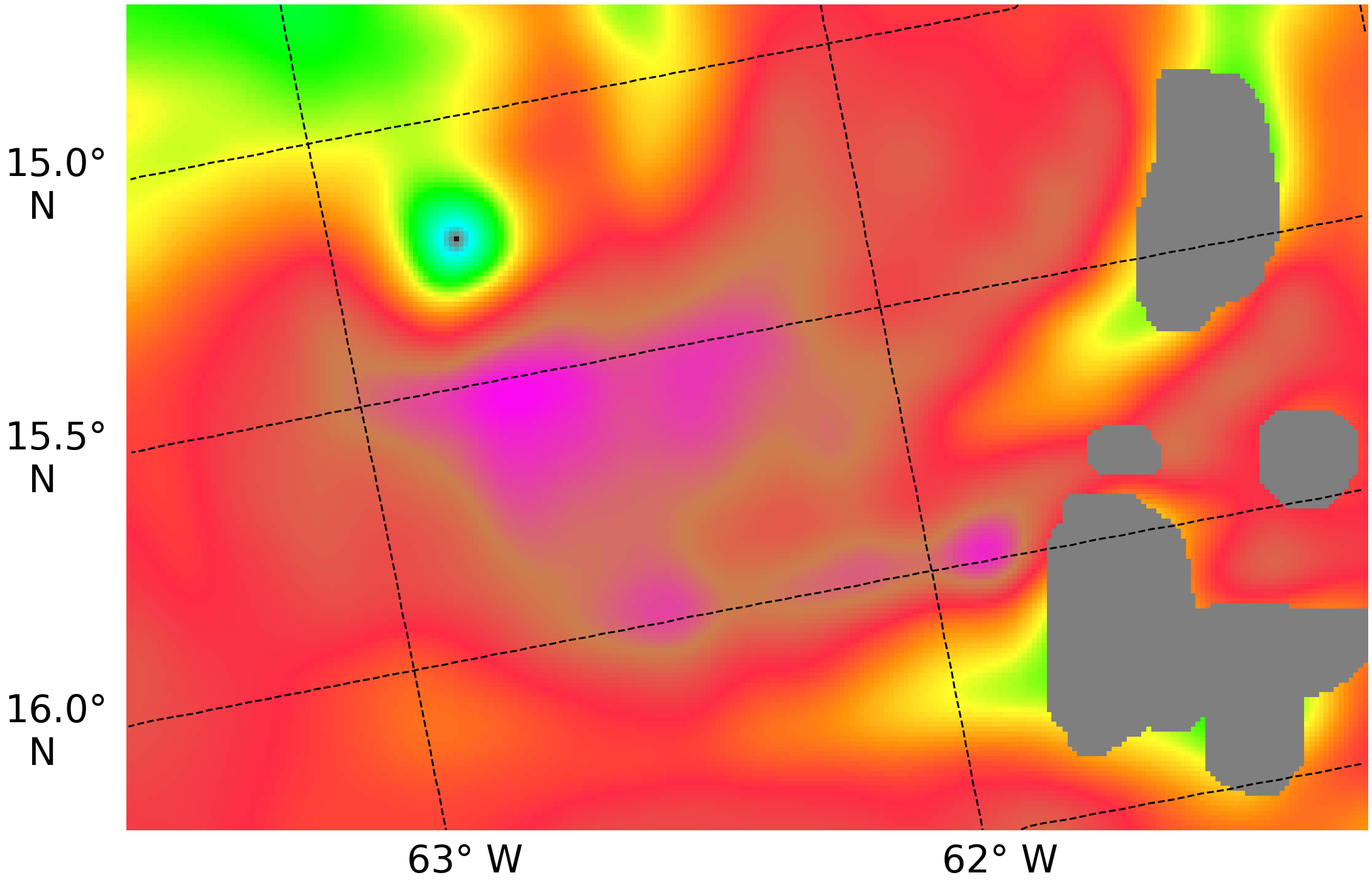}\\

    & \multicolumn{2}{c}{\includegraphics[width=0.45\linewidth]{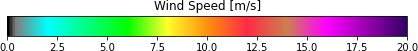}}\\
    \end{tabularx}}
    
    \caption{\textcolor{black}{Example of the discrepancies between ECMWF wind speed estimates and SAR observation. Lands are greyed out. From top to bottom: The Sea Surface Roughness from the SAR observation, the wind speed estimated from a geophysical model function and the wind speed from ECMWF's forecast atmospheric model. Red arrows indicate an increase in reflectivity caused by the rain, while green arrows indicate probable decreases.}}
    \label{fig:annexe2}
\end{figure*}

Figure \ref{fig:annexe2} illustrates some discrepancies between the SAR inversion and the ECMWF wind speed. 
\textcolor{black}{We displayed the observations displayed in this paper in satellite geometry, meaning that the satellite is observing from the left, with an incidence angle increasing along the x-axis independently from the ascending or descending passes.}
The first line, observed on May 17\textsuperscript{th} at 23:05:21, shows an area of low sea surface roughness in the lower half. The SAR inversion contradicts the ECMWF results by more than 5 m/s over most of the IW. This discrepancy is related to a time difference of 55 minutes. The second line, in which there is a time difference of 79 minutes (observations acquired on August 27\textsuperscript{th} at 22:19:26), indicates a misalignment between the ECMWF wind speed and the SAR inversion, as illustrated by the position of the eye of the cyclone. \textcolor{black}{On the other hand, the wind speed derived from the SAR observations appears to be contaminated by rain signatures. In the first observation, the rainfall causes an overestimation of the wind speed South of 33.5°N due to the ring-waves generated by the impact of hydrometeors. In the second observation, overestimation of wind speed is visible in several areas, often closely associated with underestimation. This effect is particularly noticeable around 62°W, although it is challenging to determine whether it is caused by wave damping or atmospheric attenuation. It should be noted that the wind direction used prior to GMF computation is provided by ECMWF. Therefore, the misalignment of the cyclone eye likely introduces a bias in the estimation of wind speed.}

In the following sections, all the analyses rely on the ECMWF wind speed to ensure \textcolor{black}{the} independence with the presence or absence of rain, despite occasional errors.

\subsection{SAR preprocessing}

The TOPSAR process \cite{10.1109/TGRS.2006.873853} used for the IW mode divides each observation into three subswaths along the azimuth, themselves divided into several bursts along the range. The calibration of each subswath and burst is performed by Sentinel-1's Instrument Processing Facility (IPF) by calculating the theoretical gain of the SAR antenna. 
\textcolor{black}{The resulting Normalized Cross-Section Reflectivity (NRCS) is further corrected for thermal noise, using the noise equivalent sigma nought (NESZ) annotation included in the SAR products \cite{Sentinel1ProductSpecification}. The residual NESZ is still noticeable on cross-polarization channels (i.e. VH and HV polarizations), due to a lower signal power, but not on co-polarization channels (i.e. VV and HH polarizations) \cite{10.1109/tgrs.2017.2765248}. In this study, \textcolor{black}{we only use the VV polarization}. To reduce the dependence on the incidence angle, the NRCS is divided by the NRCS corresponding to a wind speed of 10 m/s and a direction of 45° relative to the satellite heading (also called neutral wind), following the methodology used in \cite{10.1002/gdj3.73}. \textcolor{black}{We perform t}his normalization using the Geophysical Model Function CMOD5.N \cite{10.1175/2009jtecho698.1} which links the NRCS, the incidence angle, and the wind vector. The result of this normalization is called Sea Surface Roughness (SSR).}

The GMF is computed as per Equation \ref{eq:GMF} with $U$ the neutral wind speed at 10m, $\theta$ the incidence angle and $\phi$ the wind direction relatively to \textcolor{black}{azimuth (i.e. the satellite heading)}. $a$, $b$, $c_1$ and $c_2$ are parameters of the GMF.

\begin{equation}
    \sigma_0 = a(U, \theta)[1 + c_1(U, \textcolor{black}{\theta})\cos \phi + c_2(U, \theta)\cos2\phi]^b
    \label{eq:GMF}
\end{equation}

\textcolor{black}{The SSR is fed to the model on a linear scale, clipped between 0 and 6. However, the SSR is displayed in the present using a logarithmic scale, where a null value corresponds to neutral wind conditions.}

\subsection{Enhanced collocalized Sentinel-1/NEXRAD dataset}

Our collocalized dataset extends the approach introduced in \cite{rs13163155} to enhance the quality of the dataset used to train deep learning schemes. Especially, we do not require the filtering crietrion considered below to apply in inference mode. 

\textcolor{black}{Weather radars are affected by multiple factors that reduce the agreement with the collocation of the SSR from Sentinel-1}. Due to the Earth curvature \textcolor{black}{and the inclination (0.5°)}, the altitude of the observed volume increases with the distance from the ground station. As such, since the SAR imagery observes the ocean surface, both information gradually deco\textcolor{black}{r}relate with the distance to the coast. This concordance is difficult to qualify automatically since the altitude observed by the ground station depends not only on the distance but also on the refractive index of the air, which is related to the atmospheric temperature \textcolor{black}{and humidity} gradient\textcolor{black}{s}. As explained by \cite{10.1016/j.rse.2016.10.015}, the bright rain signatures in SAR may also be due to strong reflectivity from the melting layer (4th physical process aforementioned) \textcolor{black}{where snowflakes melt into droplets}. Other phenomena can \textcolor{black}{also} hinder the radar's ability to provide accurate information. Topography, for example, can mask rain signatures located behind an obstacle. The accumulation of these sources of discrepanc\textcolor{black}{ies} is difficult to quantify automatically and require manual verification.

In addition, to ensure \textcolor{black}{a consistent} SAR processing for all observations, \textcolor{black}{we focus on  SAR} observations \textcolor{black}{from} March 2018 \textcolor{black}{onward}. This corresponds to Sentinel-1 IPF version 2.90, which has improved noise correction and the associated signal-to-noise ratio (SNR). In total, these constraints reduced the number of available wide swath SAR products from 1064 to 53. 

Each of these IWs is divided into patches of approximately 25 x 25 kilometers and 256 x 256 pixels. Patches were extracted from the swaths with a step size of half their width. This extraction ensures that a \textcolor{black}{meteorological and oceanic} situation occurring at the edge of one patch is found in the center of the next patch. It also implies that each pixel is present in four patches.

\begin{figure}[!ht]
    \centering
    \centering
    \resizebox{\linewidth}{!}{%
    \begin{tabular}{ccc}
    \multirowcell{2}{SAR\\observation} & \multirowcell{2}{Original\\collocation} & \multirowcell{2}{Corrected\\collocation} \\\\
    \includegraphics[width=0.3\linewidth]{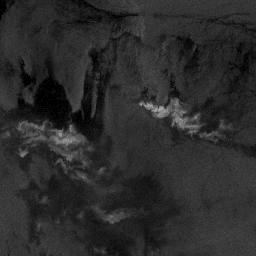}
    & \includegraphics[width=0.3\linewidth]{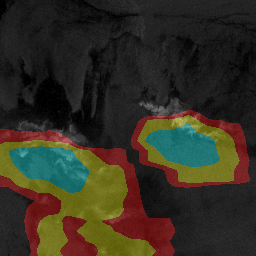}
    & \includegraphics[width=0.3\linewidth]{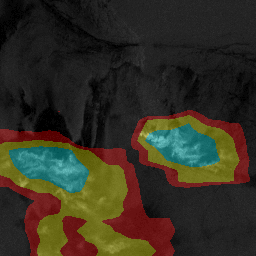}\\
    \end{tabular}}
    \includegraphics[width=0.95\linewidth]{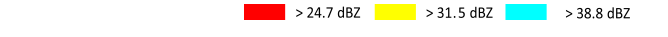}
    \caption{Realignment example. On the left, a patch (20 x 20 km) observed on May 5\textsuperscript{th} 2018 at 23:05:20. In the center, the corresponding NEXRAD measurement. The cyan-colored area does not overlap perfectly with the SAR signature (center), therefore, we perform a manual realignment (right).}
    \label{fig:registration}
\end{figure}

\textcolor{black}{We withdraw p}atches \textcolor{black}{which did not fulfil the following requirement: patches include less than 50\% of lan pixels; the maximum NEXRAD reflectivity was less than less than 25
dBZ for all patches; no patch is  rain-free.} 
After this filtering step, to maximize the overlap of SAR and weather radar signatures, the patches are aligned manually. This is carried out independently for each patch using a constant translation of the NEXRAD measurement to overlap the SAR signatures. An example of this operation is shown in Figure \ref{fig:registration}. This geographic repositioning corrects remaining collocation problems that may be related to the displacement or evolution of rain cells between NEXRAD and SAR observation times or to the different altitudes at which the phenomena were observed. 

\begin{figure*}[!ht]
    \centering
    \includegraphics[width=\linewidth]{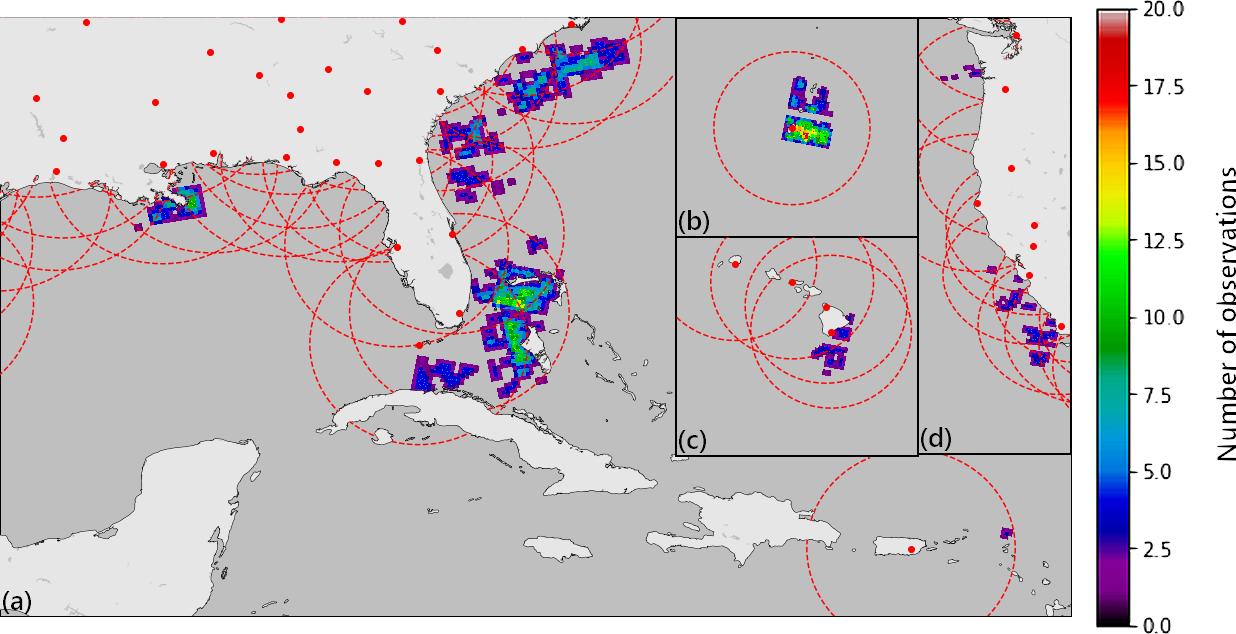}
    \caption{Geographic distribution of patches in the dataset centered around Florida (a), Guam (b), Hawaii (c), and California (d). NEXRAD stations are represented by a red dot. Circles correspond to a 350 km range. Values are the number of observations over the corresponding latitude/longitude.}
    \label{fig:dataloc}
\end{figure*}

\textcolor{black}{During the manual realignment process, we removed the patches which involved a registration displacement between Nexrad and SAR observations larger than a few kilometers. We also discarded the patches which involve the shapes of the rain signatures too dissimilar to achieve a sensible registration.  Overall, this curation process reduced the number of patches from 9574 from IWs to 1570.} \textcolor{black}{We report} patch locations, alongside the NEXRAD stations in Figure \ref{fig:dataloc}.

\textcolor{black}{The misalignment distance between Sentinel-1 and NEXRAD rain signatures is found to correlate with the distance to the NEXRAD ground station ($R^2$ = 40.4\%). This indicates that the collocations are less reliable as the distance (and the altitude of the observed volume) increases. We assume that this misalignment is caused by the horizontal drift identified by \cite{10.1080/01431169008955114} between weather radar observations and in situ rain gauges.}

\bigskip

\textcolor{black}{Given the collocated patch dataset, we create training, validation, and test sets. It is important to ensure independence between each subset to evaluate the generalization performance of deep learning schemes beyond the training dataset. Since the dataset is built with overlapping patches, adjacent patches share some pixels. Additionally, two patches extracted from the same IW observation may be affected by the same biases. To ensure independence between subsets, we perform the split at the swath level, meaning that two patches extracted from the same IW observation will be in the same subset.}

\begin{figure}[!ht]
    \centering
    \resizebox{\linewidth}{!}{%
    \footnotesize
    \begin{tabular}{cc}
        \multirowcell{3}{Sea surface roughness\\depending on\\NEXRAD's reflectivity} &
        \multirowcell{3}{Number of pixels in\\the dataset depending\\on NEXRAD's reflectivity} 
        \\ \\ \\
        \includegraphics[width=0.45\linewidth]{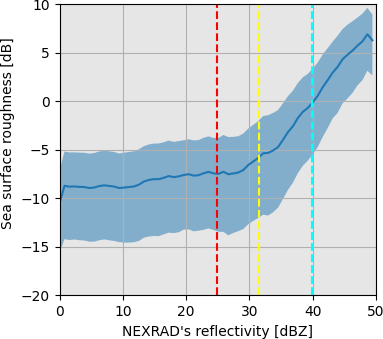} & 
        \includegraphics[width=0.45\linewidth]{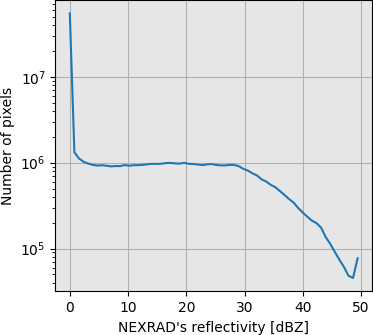} \\
        \multirowcell{3}{Sea surface roughness\\depending on\\ECMWF's wind speed} &
        \multirowcell{3}{Number of pixels in\\the dataset depending\\on ECMWF's wind speed} 
        \\ \\ \\
        \includegraphics[width=0.45\linewidth]{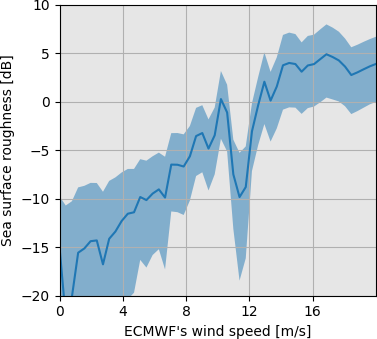} & 
        \includegraphics[width=0.45\linewidth]{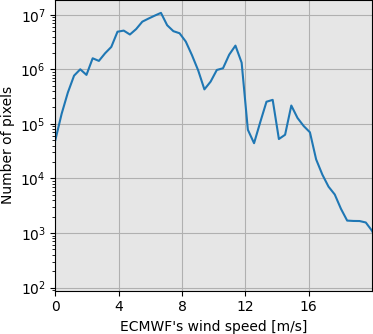} \\
    \end{tabular}}
    \caption{Evolution of sea surface roughness (and associated number of pixels) as a function of NEXRAD reflectivity or ECMWF wind speed and their associated number of pixels. The red, yellow and cyan vertical lines are the threshold values used to separate the four precipitation classes. The blue area represents the standard deviation around the mean at each point. The decrease in sea surface roughness at 11 m/s is due to a single IW (taken on May 17\textsuperscript{th} 2018 at 23:05:21). Comparisons between SAR-derived and ECMWF's wind speed confirm E\textcolor{black}{CM}WF is over-estimating the wind speed for this particular case.}
    \label{fig:a_priori_distribution}
\end{figure}

\textcolor{black}{
To ensure a similar distribution of rainfall in each data subset, the subsets were balanced on both the NEXRAD reflectivity and the wind speed, which presumably has an impact on rainfall prediction capabilities.} This assumption is driven by the known increase in sea surface roughness under the impact of rain and wind, as illustrated in Figure \ref{fig:a_priori_distribution}. The NEXRAD reflectivity and Sentinel-1 backscatter \textcolor{black}{also} decorrelate for reflectivities below 30 dBZ.

The resulting distributions for each subset of the data are detailed in Table \ref{tab:dataset_distribution}. As noted, the dataset suffers a lack of data at higher reflectivity and wind speeds. Indeed, as an example, only two wide-swaths contain wind speeds above 12 m/s, and only one has wind speeds above 16 m/s. It can also be noted that the standard deviation of the SAR surface roughness is higher when wind speed is higher than 12 m/s. 

\begin{table}[!ht]
    \small
    \centering
    \resizebox{\linewidth}{!}{%
    \begin{tabular}{|cc||c|c|c||c|}\hline
         \multicolumn{2}{|c||}{\multirowcell{2}{Dataset}}
             & \multirowcell{2}{Train\\(39 IW)}
             & \multirowcell{2}{Validation\\(7 IW)}
             & \multirowcell{2}{Test\\(7 IW)}
             & \multirowcell{2}{\% of\\the total}
         \\&&&&&\\ \hline
        \multirowcell{4}{\rotatebox[origin=c]{90}{\footnotesize \parbox{1.2cm}{\centering NEXRAD reflectivity [dBZ]}}}
        &${[0, 24.7[}$
            & 79.5 \% 
            & 9.6 \% 
            & 10.9 \% 
            & 85.1 \% \\
         &${[24.7, 31.5[}$
            & 79.9 \% 
            & 9.6 \% 
            & 10.5 \% 
            & 7.7 \%  \\
         &${[31.5, 38.8[}$
            & 79.3 \% 
            & 9.7 \% 
            & 11.0 \% 
            & 5.4 \% \\
         &$\geq 38.8$
            & 79.0 \% 
            & 9.8 \% 
            & 11.2 \% 
            & 1.8 \%  \\ \hline
        \multirowcell{5}{\rotatebox[origin=c]{90}{\footnotesize \parbox{1.6cm}{\centering ECMWF wind speed [m/s]]}}}
        &${[0, 4[}$
            & 79.3 \% 
            & 9.7 \% 
            & 11.0 \% 
            & 11.7 \%   \\
        &${[4, 8[}$
            & 79.1 \% 
            & 9.7 \% 
            & 11.1 \% 
            & 69.7 \%   \\
        &${[8, 12[}$
            & 79.1 \% 
            & 9.5 \% 
            & 11.3 \% 
            & 17.1 \%   \\
        &${[12, 16[}$
            & 100 \% 
            & 0.0 \% 
            & 0.0 \% 
            & 1.5 \%   \\
        &$\geq 16$
            & 100 \% 
            & 0.0 \% 
            & 0.0 \% 
            & 0.1 \%   \\ \hline
    \end{tabular}}
    \caption{Per-pixel distribution of NEXRAD reflectivity and ECMWF wind speed for the patches contained in each set.}
    \label{tab:dataset_distribution}
    \vspace{-0.5cm}
\end{table}

The reflectivity is divided into four intervals: [0, 24.7], [24.7, 31.5], [31.5, 38.8] and [38.8, +$\infty$] dBZ. Incidentally, according to the general NEXRAD radar formula \cite{NexradHandbook}, these intervals can be approximated in terms of rainfall by thresholds at 1 mm/h, 3 mm/h and 10 mm/h. The precipitation estimation is tackled not from a continuous regression, which is difficult because of the aforementioned discrepancies between Sentinel-1 and NEXRAD sensors, but as the segmentation of the following \textcolor{black}{rainfall} classes: $\geq$ 24.7 dBZ, $\geq$ 31.5 dBZ, and $\geq$ 38.8 dBZ. Reflectivities below 24.7 dBZ are considered to be rain-free.
\textcolor{black}{The} direct prediction of reflectivity \textcolor{black}{values} proved difficult due to the scarcity of the strongest rain events and the \textcolor{black}{low} correlation between the reflectivity value and the SAR signature. The non-uniform distribution of NEXRAD reflectivity is also \textcolor{black}{an issue} because low reflectivities are over-represented, as shown in Figure \ref{fig:a_priori_distribution}.b. \textcolor{black}{The proposed segmentation-based formulation addresses these challenges. Overall, t}he \textcolor{black}{resulting datatset of} 1570 patches, divided in training \textcolor{black}{(1243 patches)}, validation \textcolor{black}{(153 patches)} and test sets \textcolor{black}{(174 patches)} is available on kaggle at:  \url{www.kaggle.com/rignak/sentinel1-nexrad}.

\section{Proposed Deep learning framework}

This section introduces the proposed deep learning schemes. Within a supervised training framework, we explore two neural network architectures: the first one derived from the Koch's filters \cite{10.1109/tgrs.2003.818811}, the second one based on a U-Net architecture \cite{RFB15a}.

\subsection{Koch's filter-based architecture}

Koch's filters, introduced by \cite{10.1109/tgrs.2003.818811}, are four different high-pass filters that each detect different patterns thus allowing the detection  of heterogeneous areas of ocean surface roughness. Their original use was to identify areas where backscatter is caused by non-wind phenomena (ships, rain, interference, tidal currents...), as this would exclude these areas from a wind speed/direction estimate. Koch's filters can be optimized to produce binary rainfall maps, as precipitation is a major source of heterogeneity \cite{rs13163155}.

\begin{figure*}[!ht]
    \centering
    \includegraphics[width=0.8\linewidth]{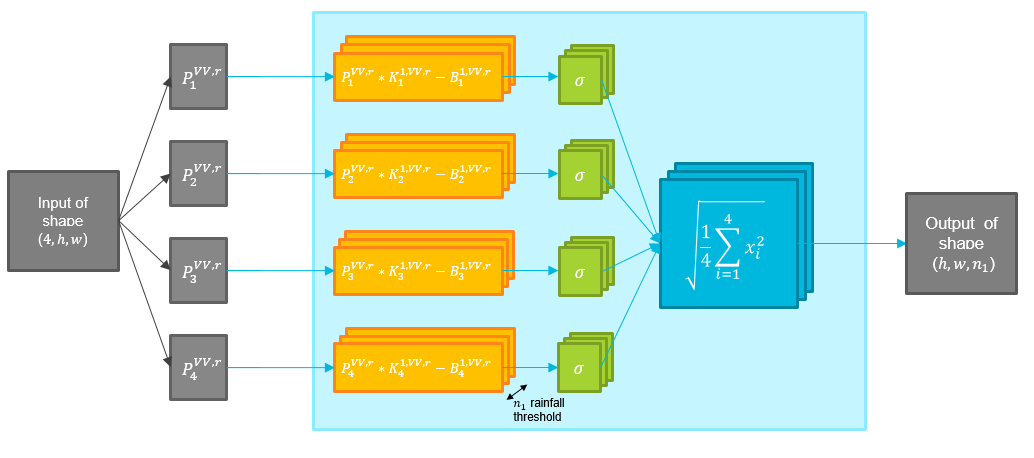}
    \caption{Architecture of the multi-threshold Koch's filters as a convolutionnal neural network. $P_i^{VV,r}$ is the output of the high-pass filter $i$ at the resolution $r$, for the $VV$ polarization. $\sigma$ is the sigmoid function defined as $\sigma(x) = [1+ \exp(-4(x+0.5))]^{-1}$. $K_{j}^{i, VV, r}$ and $B_{j}^{i, VV, r}$ are the scaling function parameters for each resolution $r$, polarization $VV$, filter $j$ and precipitation regime $i$. The results are fused along the filters by a quadratic mean.}
    \label{fig:koch_cnn}
\end{figure*}

Specifically, Koch's filters are defined as four different high-pass filters scaled by a linear function and clipped to maintain the result between 0 and 1. The output of the filters is the root mean square of these clippings. \cite{rs13163155} estimated thresholds in order to derive binary rain maps from this final value, depending on the resolution and polarization of the input. We extended this framework to multi-threshold segmentation by rewriting the Koch's filters as a Convolutional Neural Network (CNN) defining the scaling function parameters. The four high-pass filters were used on the input and left side as in the original version. To guarantee a non-zero gradient, the clipping is replaced by the sigmoid function $\sigma(x) = [1+ \exp(-a(x+b))]^{-1}$. We set $a=4$ and $b=0.5$ so that the inflection point is at x=0.5, $\sigma(0.5)=0.5$ and $\frac{d\sigma}{dx}(0.5)=1$. The change in activation affects the filter result, but the relative difference from the original Koch's filters is only 0.8\% when initialized with the same parameters. Figure \ref{fig:koch_cnn} illustrates this Koch's filters incorporated into the CNN. This formulation allows for segmentations for different rain thresholds, unlike the original rain detection \cite{rs13163155}. 

The model is trained to minimize the mean square error with the \textcolor{black}{ADAptive Moment estimation (ADAM)} optimizer, a learning rate of $10^{-3}$ over 200 epochs with a batch size of 32. \textcolor{black}{ADAM is an optimizer commonly used in deep learning schemes as it combines gradient momentum and adaptive learning rate to accelerate the convergence of the model \cite{kingma2017adam}}. As previously mentioned, the convolution kernels were initialized following the original Koch's filters formulation \cite{10.1109/tgrs.2003.818811}. 

\subsection{U-Net architecture}

Among the variety of state-of-the-art neural architectures for image segmentation and image-to-image translation problems, we consider here a U-Net architecture \cite{RFB15a}. \textcolor{black}{U-Net is an auto-encoder model, which is a neural network composed of an encoder that projects the input into a latent space, usually of smaller dimension, , and a decoder that, on the contrary, projects latent maps back into the original image space. U-Net uses "skip connections" to propagate intermediate activation maps between the encoder and the decoder, bypassing the central part of the network and facilitating the preservation of details.} This architecture is well established and has already been used in SAR imagery for sea ice concentration estimation \cite{10.1109/JSTARS.2021.3074068} and semantic segmentation \cite{Colin2022}. 

\begin{figure*}[!ht]
    \centering
    \includegraphics[width=0.8\linewidth]{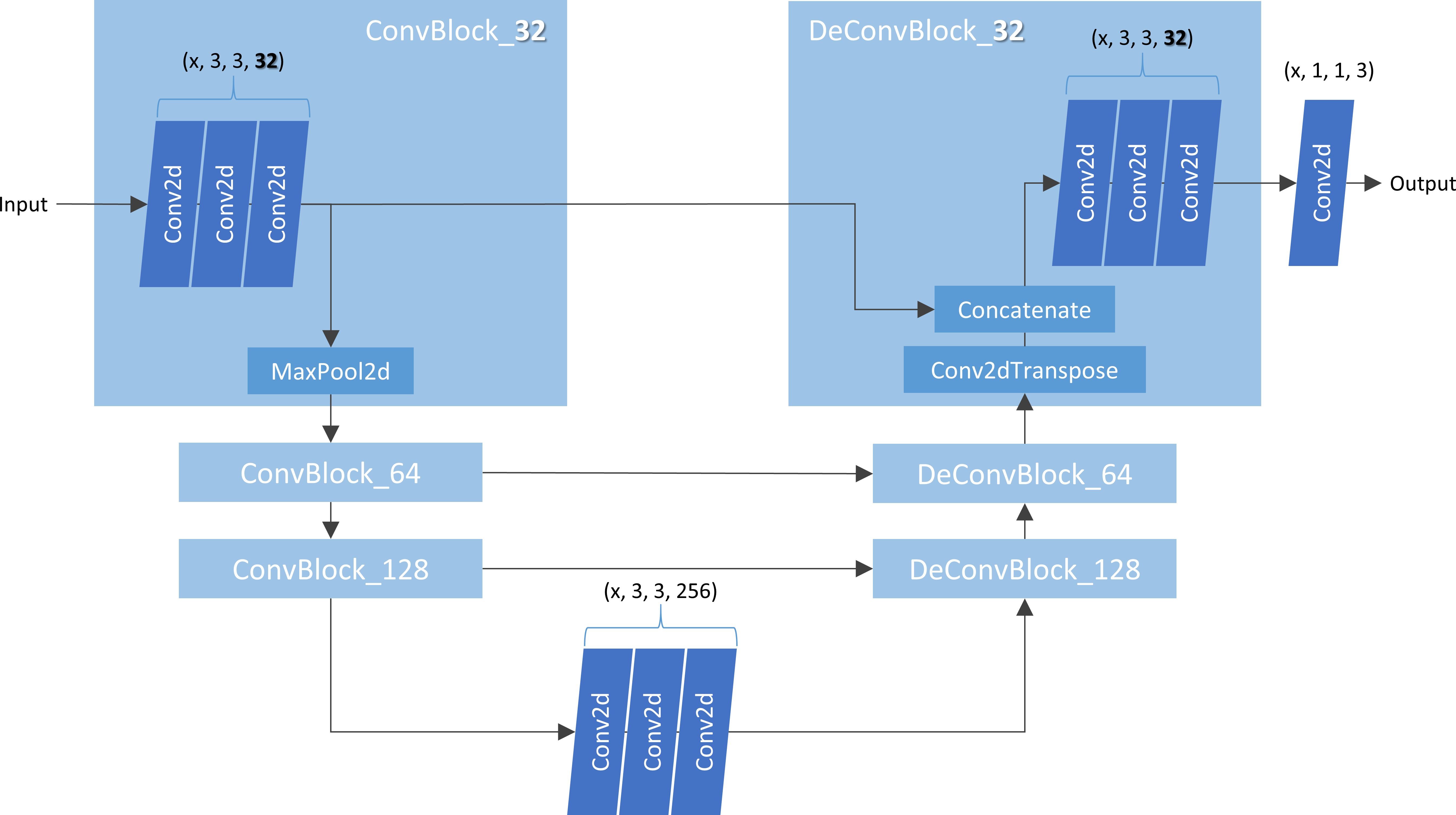}
    \caption{Architecture of the U-Net model used to classify rainfall amount (from light to heavy) using Sentinel-1 ocean surface roughness.}
    \label{fig:unet}
\end{figure*}

The specific model used is shown in Figure \ref{fig:unet}. Compared to the original U-Net model, it has one less stage to reduce the receptive field and ensure that, when applied to full IW observations divided into overlapping tiles, the output mosaic has continuity between adjacent tiles. The width of the theoretical receptive field is 140 pixels, but the effective receptive field, which is smaller due to the contribution of neighboring pixels that exponentially decreases with distance \cite{1701.04128}, is small enough to ensure continuity. The number of weights, independent of input size and spatial resolution, was 3,117,731.

\begin{figure*}
    \includegraphics[width=\linewidth]{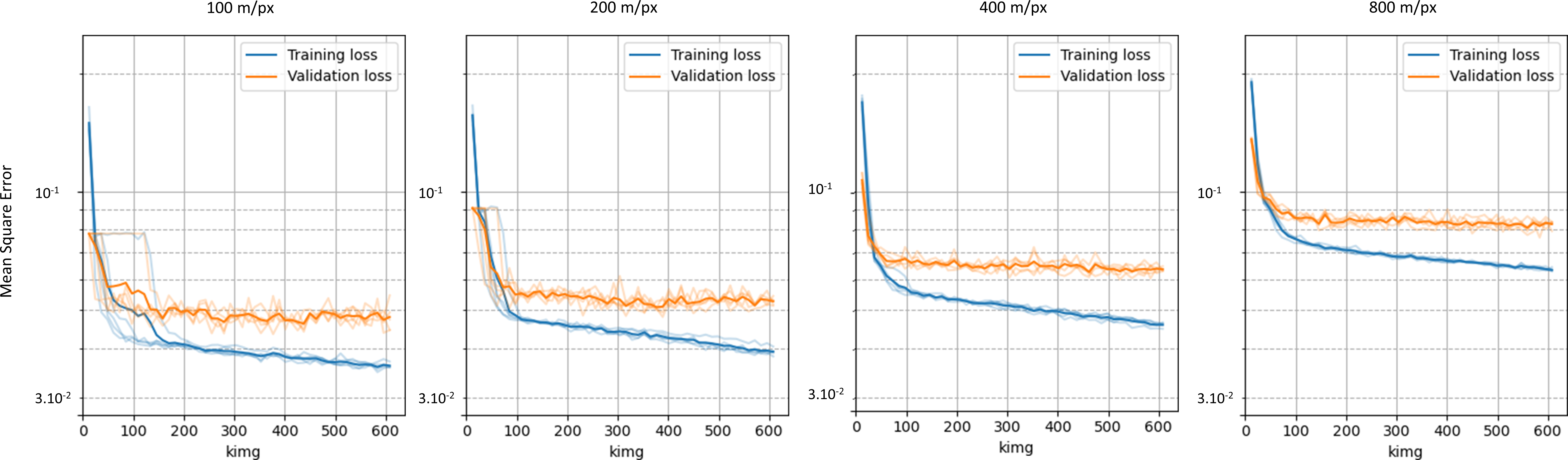}
    \caption{\textcolor{black}{The evolution of the Mean Square Error during training for the U-Net models and both the validation and training sets. Each faded curve corresponds to a single training, while the mean over the five trainings is highlighted. One epoch corresponds to 1243 images.}}
    \label{fig:history}
\end{figure*}

The model was trained to minimize the mean square error, using ADAM with a learning rate of $10^{-5}$, for 500 epochs. In all experiments, 500 epochs were sufficient to achieve loss convergence. The batch size was 32, except at 100 m/px where GPU memory constraints led to reducing the batch size to 16. \textcolor{black}{Batch size refers to the number of samples on which the gradients are evaluated at each step of the optimization process. Decreasing the batch size reduces memory usage but may introduce instability in the training process.}. The code used to train the model is accessible at \url{https://github.com/CIA-Oceanix/SAR-Segmentation/tree/oceanix}. \textcolor{black}{The evolution of the training and validation losses during the training is presented in \ref{fig:history}.} 

\section{Evaluation framework}

Existing Koch's filters are designed to produce binary maps of rain presence or absence. To compare this framework with our multi-\textcolor{black}{class} models (i.e. the fine-tuned Koch's filters and U-Net), we computed the F$_1$-score on the binary segmentation problem for each threshold (24.7, 31.5 and 38.8 dBZ). The F$_1$-score is defined as the harmonic mean of recall and precision. Recall is the average diagonal value of the row-normalized confusion matrix. It is also known as the producer's accuracy. Precision is the diagonal mean value of the column-normalized confusion matrix. It is also known as the user's accuracy. 
Denoting $TP$ the True Positive rate, $FP$ the False Positive rate and $FN$ the False Positive rate

\begin{align}
    Precision = \frac{TP}{TP + FP} & ; & Recall = \frac{TP}{TP + FN} \nonumber
\end{align}
    
When evaluated on a binary segmentation problem (through a 2 x 2 confusion matrix), we call it the "Binary F$_1$-score". This value indicates the ability to separate rain-free \textcolor{black}{patches} from rain\textcolor{black}{y ones}. The F$_1$-score is also used to evaluate the ability to distinguish between different \textcolor{black}{reflectivity bins}. In this case, it is computed by taking the mean precision and recall. This F$_1$-score is indicated as the "Multiclass F$_1$-score". F$_1$-score being the harmonic mean of precision and recall, it has the advantage of being resilient to data imbalance, which is particularly important for stronger \textcolor{black}{rainfall} events less likely to appear on the samples.
 
We therefore compare the \textit{Binary Koch's filter}, which is the baseline and state-of-the-art in rain detection, to the \textit{Fine-tuned Koch's filter} (the CNN-embedded multi-label Koch's filter), and the \textit{U-Net} architecture. For the latter, results using a dataset without the manual realignment are also provided to justify the need for this particular operation. 
The U-Net models were trained and tested at resolutions of 100 m/px to 800 m/px. Because the receptive field of the Koch's filters is smaller, they were only used down to 200 m/px, in accordance with \cite{rs13163155}. Since some parts of the methodology were stochastic, such as the order of the images provided to the network or the initialization of its weights, the results are given as the mean and standard deviation over five training runs, in accordance  with \cite{DBLP:journals/corr/abs-1206-5533}. 

\section{Results and discussion}

\begin{table*}[!t]
    \footnotesize
    \centering
    \begin{tabularx}{1.\textwidth}{|Y|Y||YYY|Y|}\hline
    \textbf{Model} & \textbf{Input resolution} & \textbf{Binary F$_1$-score ($>$ 24.7 dBZ)} & \textbf{Binary F$_1$-score ($>$ 31.5 dBZ)} & \textbf{Binary F$_1$-score ($>$ 38.8 dBZ)} & \textbf{Multiclass F$_1$-score} \\\hline
    \multirowcell{3}{Binary\\Koch's filter}
        & 200 m/px & 44.3\% & 34.7\% & 22.8\% & N/A \\
        & 400 m/px & 37.3\% & 26.5\% & 15.1\% & N/A \\
        & 800 m/px & 32.9\% & 22.2\% & 11.1\% & N/A \\\hline
    \multirowcell{3}{Fine-tuned\\Koch's filter}
        & 200 m/px & 45.9\% (0.04\%) & 41.6\%( 0.06\%) & 38.7\% (2.09\%) & 34.8\% (0.2\%) \\
        & 400 m/px & 43.2\% (0.15\%) & 40.9\% (0.14\%) & 37.9\% (0.58\%) & 35.9\% (0.3\%)\\
        & 800 m/px & 38.3\% (0.05\%) & 37.2\% (0.18\%) & 32.3\% (1.65\%) & 35.2\% (0\%) \\\hline
    \multirowcell{4}{U-Net}
        & 100 m/px & 53.7\% (2.36\%) & 52.5\% (2.03\%) & 55.6\% (2.30\%) & 47.2\% (1.9\%) \\
        & 200 m/px & 50.5\% (1.69\%) & 47.5\% (1.72\%) & 48.0\% (1.87\%) & 46.0\% (3.0\%) \\
        & 400 m/px & 51.2\% (1.72\%) & 46.8\% (1.75\%) & 47.2\% (2.14\%) & \textbf{50.5\% (2.8\%)}\\
        & 800 m/px & 45.4\%(0.93\%) & 40.4\%(1.26\%) & 40.2\%(1.56\%) & 47.1\% (0.9\%) \\ \hline
    \multirowcell{4}{U-Net\\without\\realignment}
        & 100 m/px & 51.6\% (0.56\%) & 50.2\% (0.42\%) & 25.8 \% (14.76\%) & 41.0\% (1.2\%) \\
        & 200 m/px & 50.1\% (0.54\%) & 48.2\% (1.32\%) & 42.7 \% (9.24\%) & 36.1\% (2.4\%) \\
        & 400 m/px & 49.1\% (0.93\%) & 47.6\% (0.87\%) & 48.5 \% (1.35\%) & 41.0\% (1.2\%) \\
        & 800 m/px & 44.2\% (2.57\%) & 42.5\% (3.06\%) & 43.8 \% (2.41\%) & 41.5\% (1.0\%) \\ \hline
    \end{tabularx}
    \caption{Evaluation of the binary and fine-tuned Koch's filters and U-Net model on the test subset. Results are provided as a mean with standard deviation over five runs.}
    \label{tab:architecture_comparison}
    \vspace{-0.5cm}
\end{table*}

Table \ref{tab:architecture_comparison} compares the binary Koch's filters, the fine-tuned multi-label Koch's filters, and the U-Net architectures for binary segmentation (for different precipitation thresholds) and multi-class segmentation. The binary Koch's filters performs worse on the binary F$_1$-score at each precipitation threshold than do both the fine-tuned Koch's filters and the U-Net architecture. The best binary segmentation is obtained at 200 m/px for each method, and the U-Net architecture outperforms both variants of the Koch's filters. Great variability is observed in the results and can be explained by the difference in the number of parameters (24 for the finely tuned Koch's filters and over \textcolor{black}{3M} for the U-Net model).

As the multi-class F$_1$-score is not only influenced by its ability to detect precipitation but also by its ability to distinguish the severity of precipitation, it indicates higher performance at 400 m/px. Interestingly, this is also the best resolution obtained by \cite{rs13163155}, although the results were computed on a different data set. This leads one to believe that the increase in resolution, while giving more accurate information, is counterbalanced by the decrease in contextual information. Since the architecture of the network does not change, the receptive field is the same if measured in pixels, but is reduced if we consider the area covered in km². The Koch's filters are less affected by the change of context because their effective field is defined by the low pass filters they use as input.

\begin{figure}[!ht]
    \centering
    \begin{tabular}{c}
    Fine-tuned multi-class Koch's filter
         \\ \includegraphics[width=0.90\linewidth]{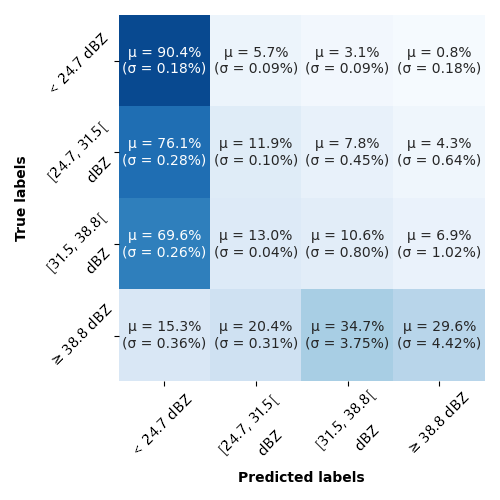}
         \\
    U-Net model with realignment
         \\ \includegraphics[width=0.90\linewidth]{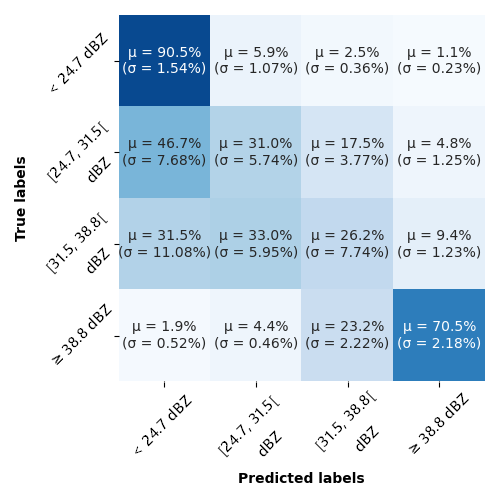}
         \\
    \end{tabular}
    \caption{Normalized confusion matrices of the fine-tuned multi-class Koch's filters (top), the U-Net model with realignment, both at 400 m/px. Result are given as mean and standard deviation over 5 training, as to mitigate the random initialization. Large standard deviations indicates unstable training.}
    \label{fig:confusion}
\end{figure}

The confusion matrices, shown in Figure \ref{fig:confusion}, indicate that the U-Net architecture (bottom) is more accurate than the fine-tuned Koch's filters (left) for each threshold (11.9\% vs. 31.0\%, 10.6\% vs. 26.2\%, and 29.6\% vs. 70.5\% for the 24.7, 31.5, and 38.8 dBZ thresholds, respectively). However, 31\% of the [31.5, 38.8[ dBZ class remains unrecognized by the model as being rain. The U-Net performs particularly well in detecting heavy rainfall, as 93.7\% of rainfall above 38.8 dBZ was predicted to be above 31.5 dBZ. The refined Koch's filters only achieved 64.4\%.

\begin{figure*}[!ht]
    \small
    \centering
    \resizebox{\linewidth}{!}{%
    \begin{tabularx}{\linewidth}{YY}
        SAR observation & NEXRAD reflectivity (groundtruth) \\
        \includegraphics[width=0.8\linewidth]{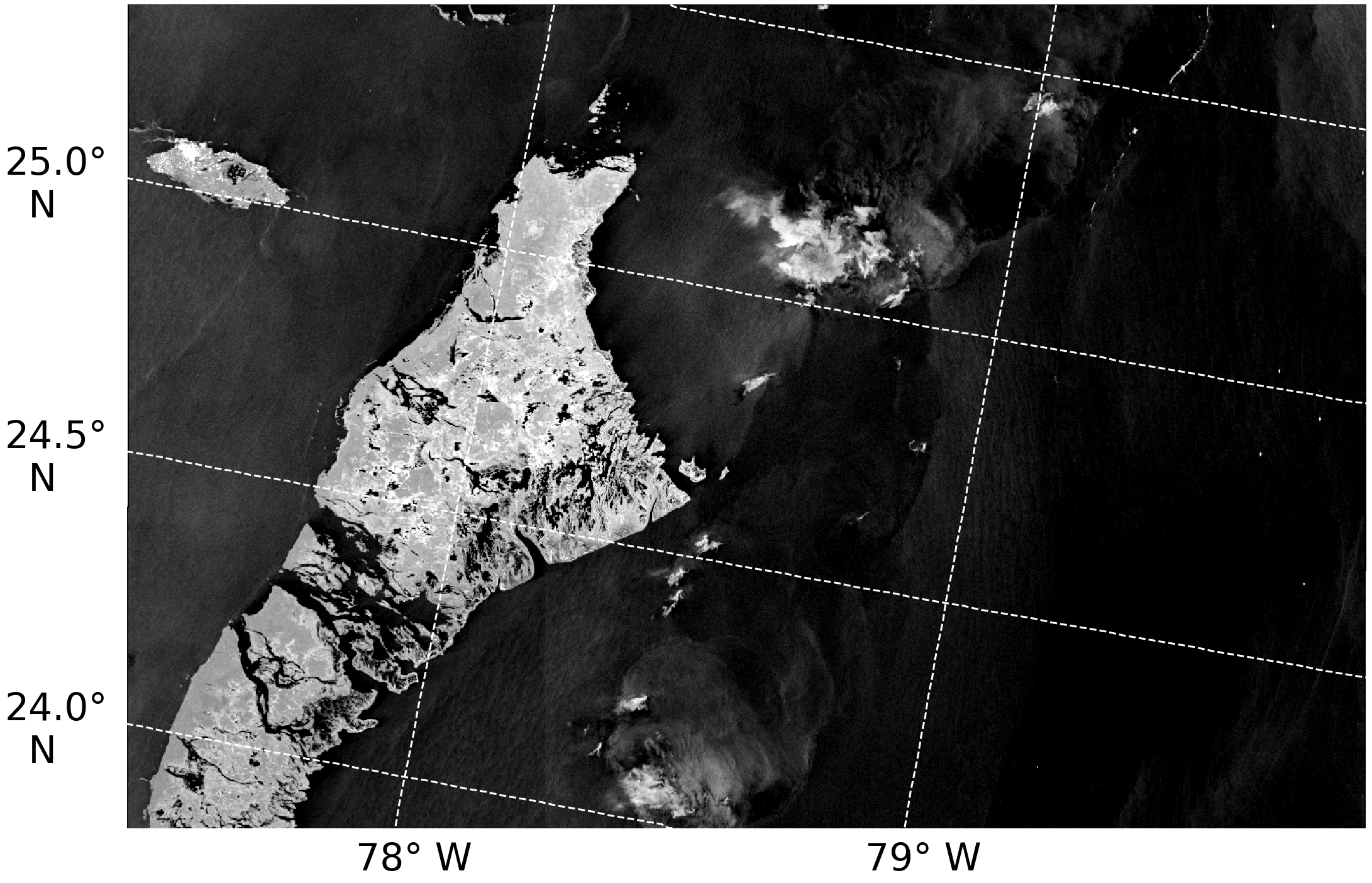} & 
        \includegraphics[width=0.8\linewidth]{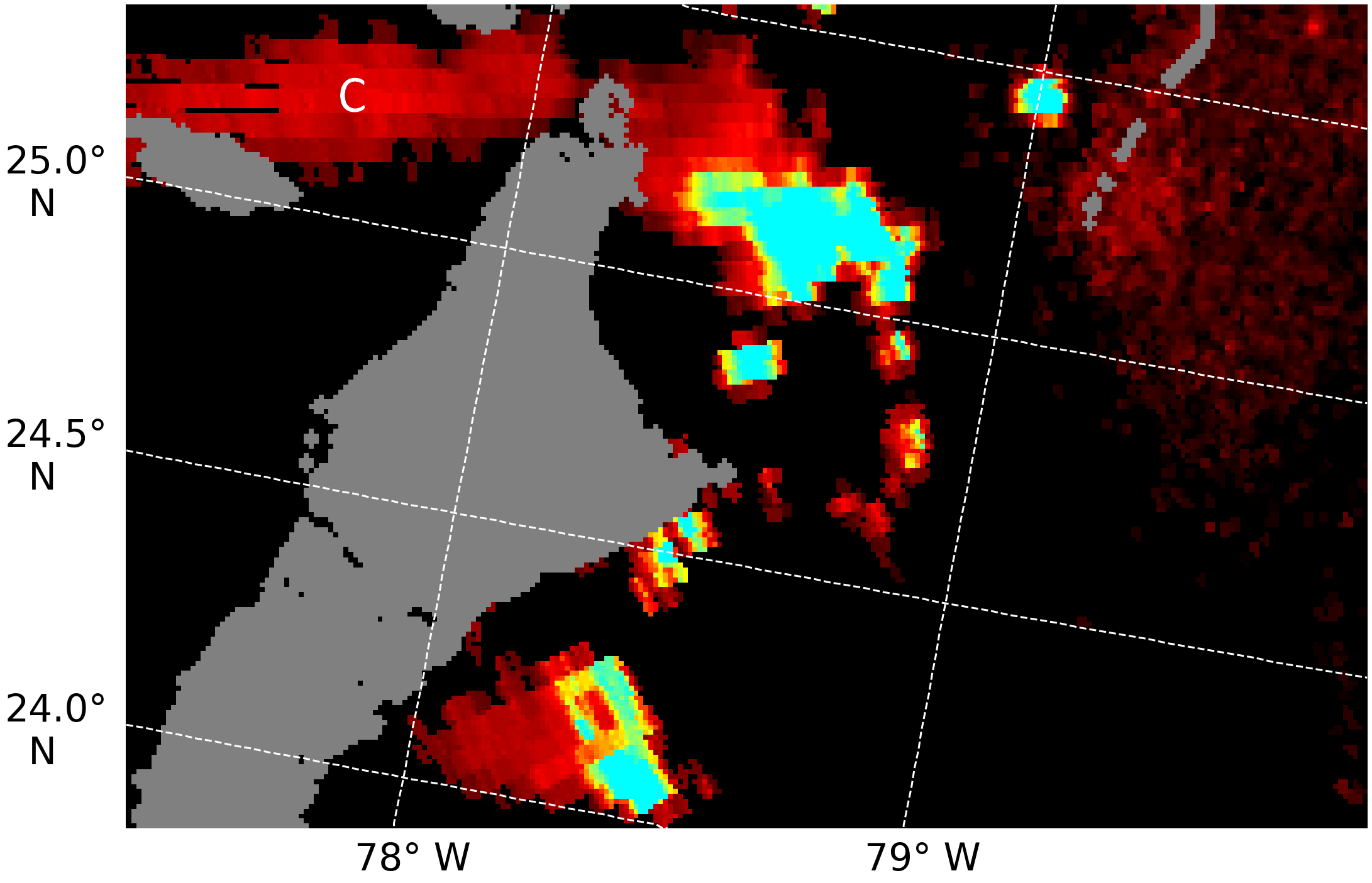}\\
        
        \includegraphics[width=0.8\linewidth]{images/examples/cmap_ssr.png} & 
        \includegraphics[width=0.8\linewidth]{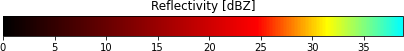}\\
        
        Prediction from Koch's filters & Prediction from U-Net model \\
        \includegraphics[width=0.8\linewidth]{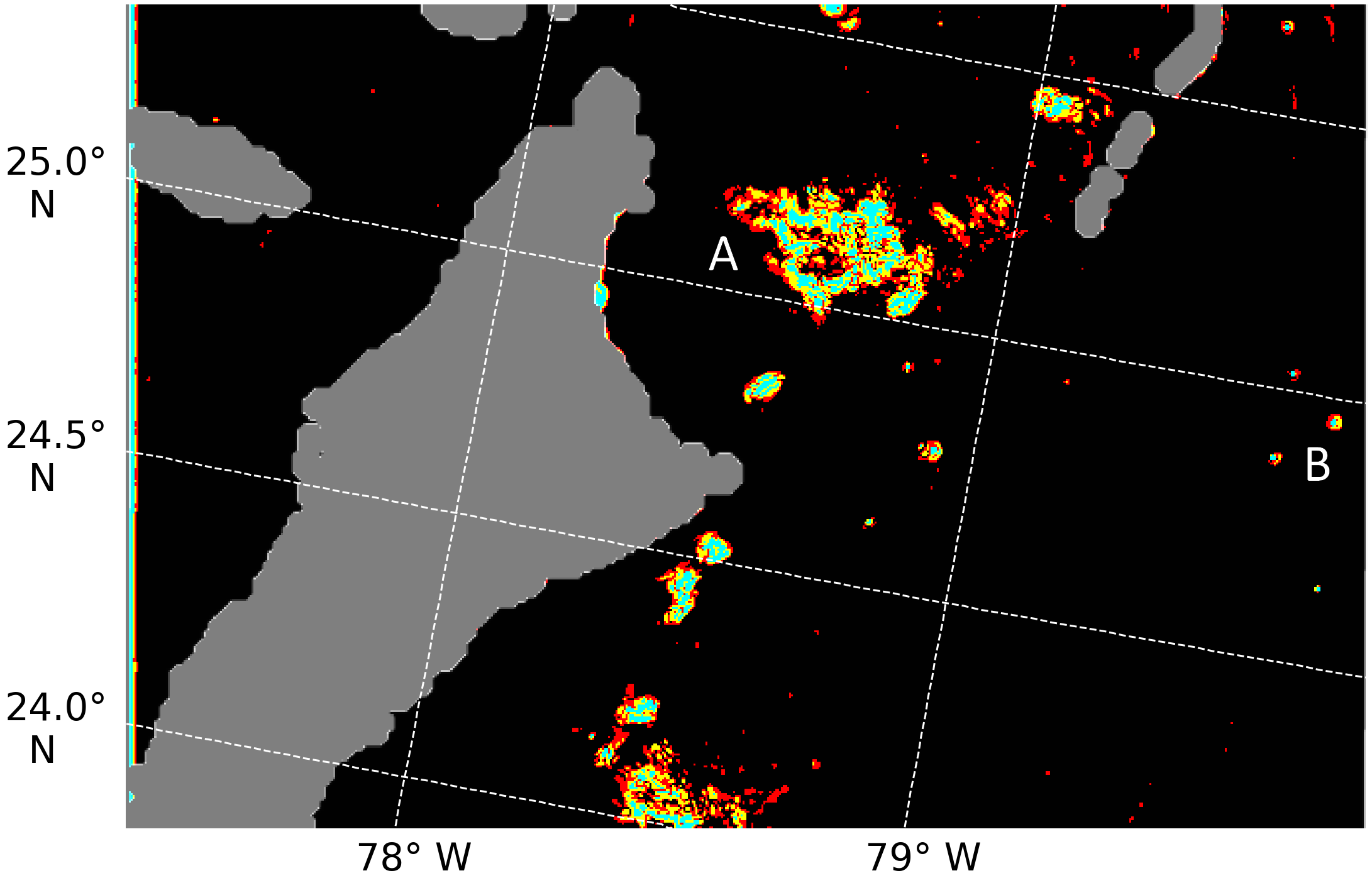} & 
        \includegraphics[width=0.8\linewidth]{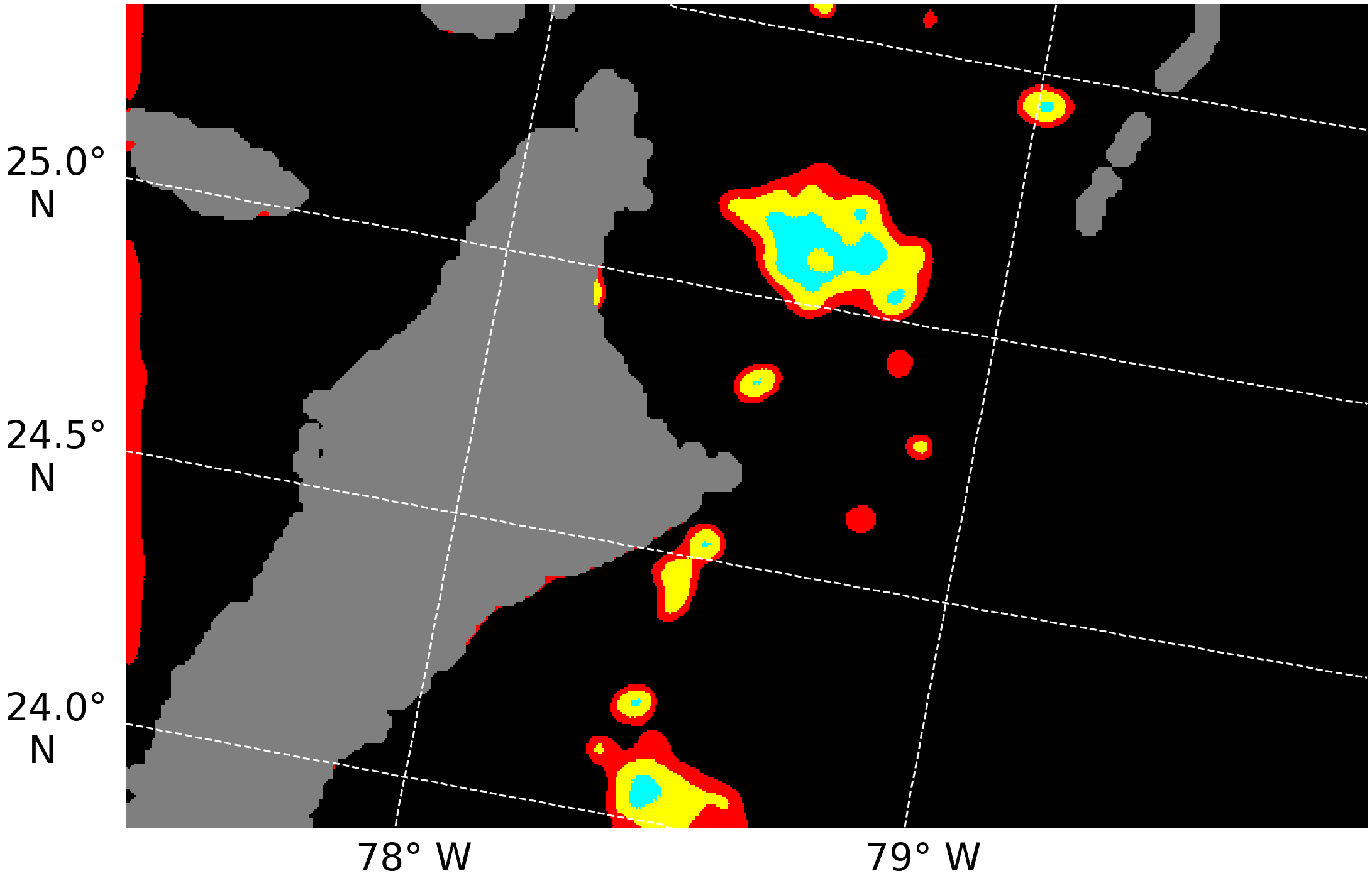}\\
        
        GOES16/ABI, band 14 (11.2 µm) & \\
        \includegraphics[width=0.8\linewidth]{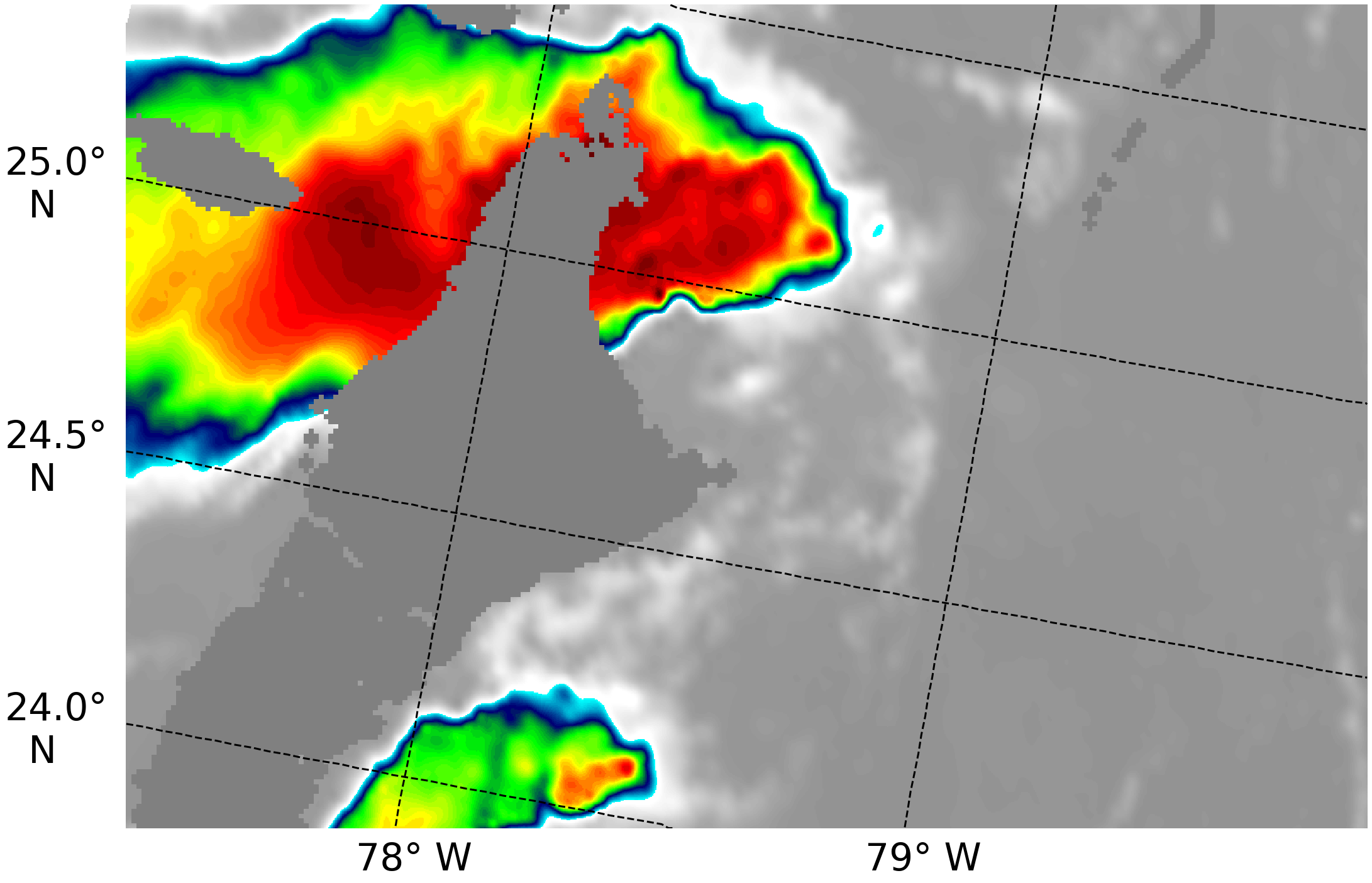} & \\
        
        \includegraphics[width=0.8\linewidth]{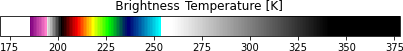} & \\
    \end{tabularx}}
    \caption{\textcolor{black}{Example of SAR-derived reflectivity estimation from the observation from April 24\textsuperscript{th} 2018 at 11:10:12.}}
    \label{fig:NEXRAD_examples1}
\end{figure*}

\begin{figure*}[!ht]
    \small
    \centering
    \resizebox{\linewidth}{!}{%
    \begin{tabularx}{\linewidth}{YY}
        SAR observation & NEXRAD reflectivity (groundtruth) \\
        \includegraphics[width=0.8\linewidth]{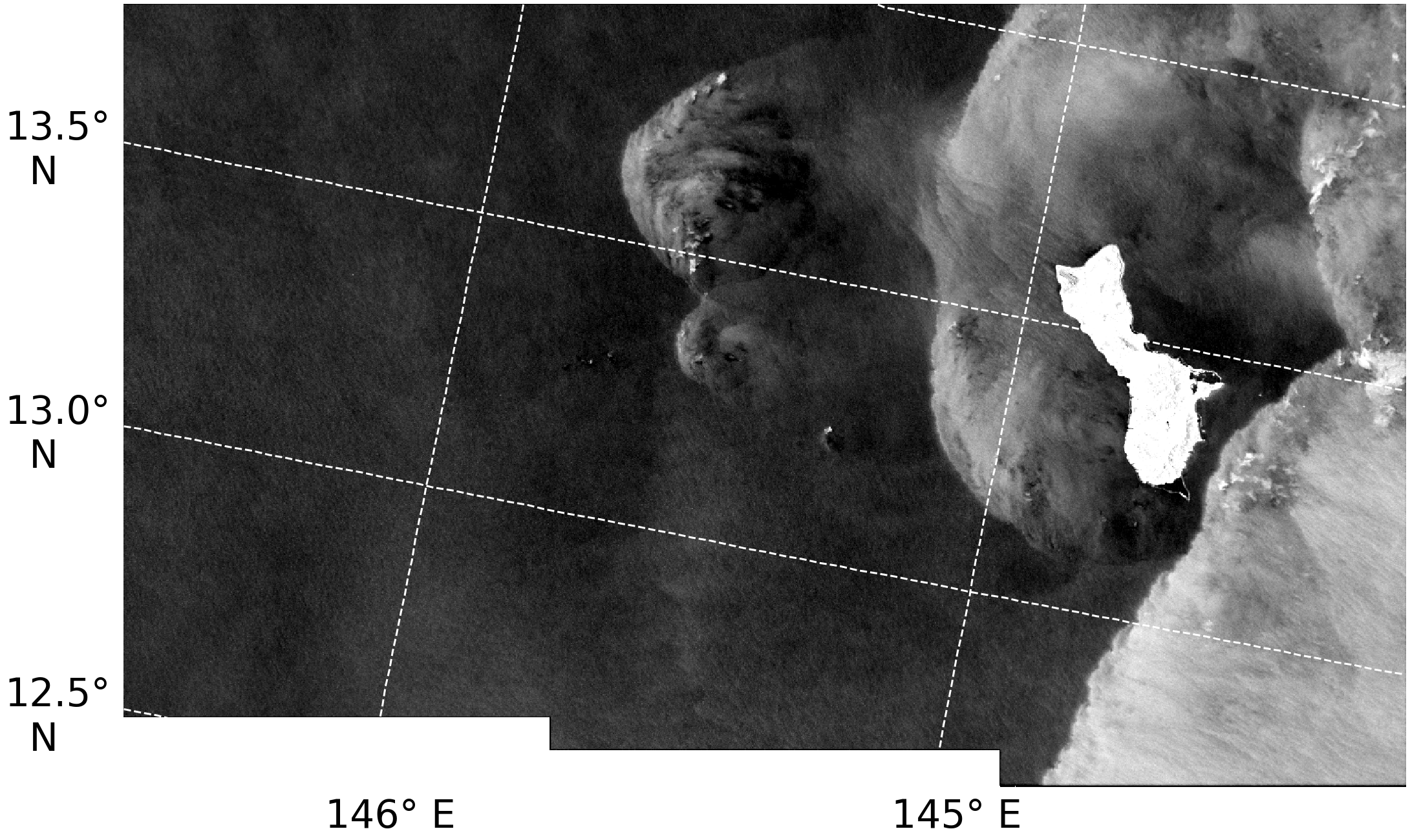} & 
        \includegraphics[width=0.8\linewidth]{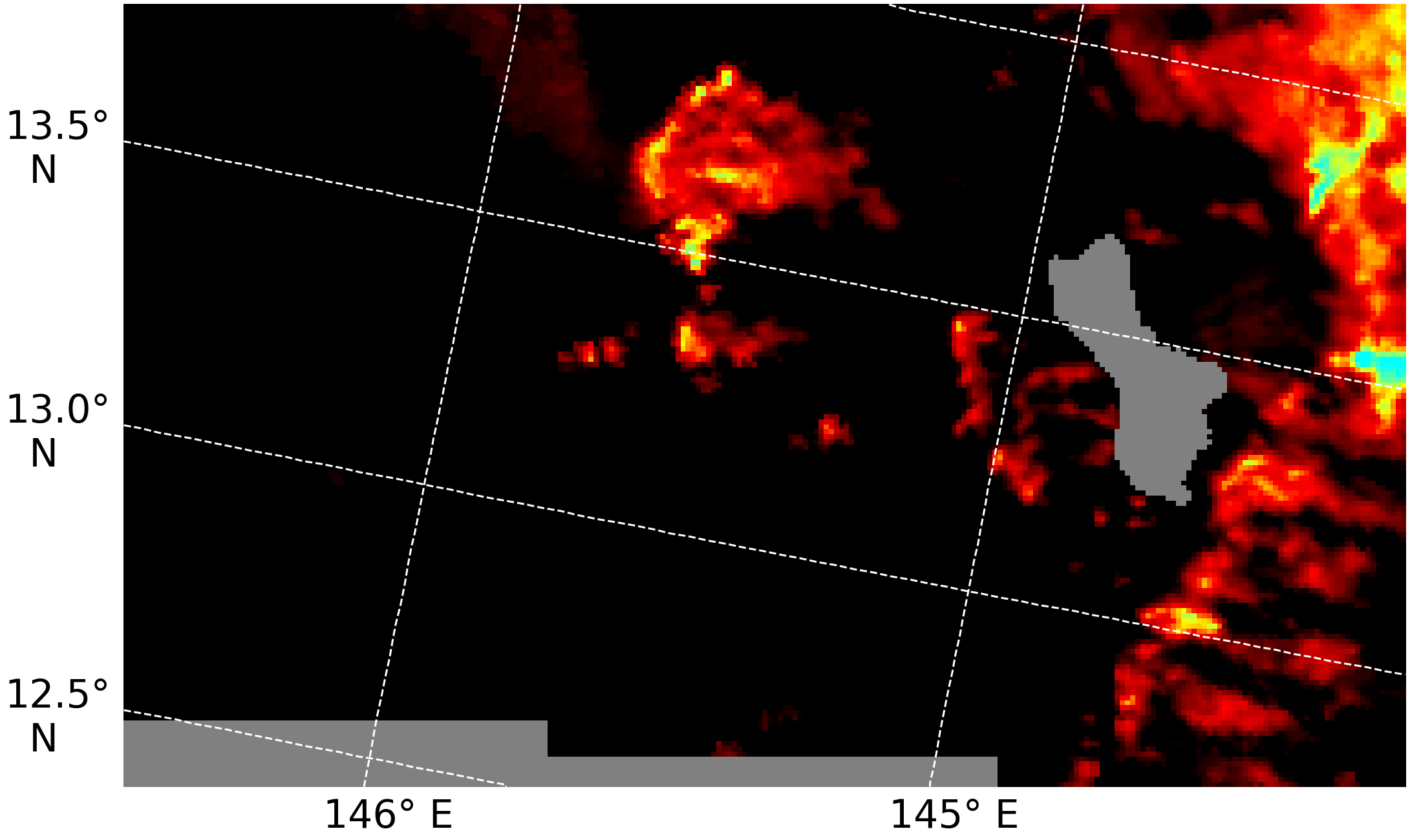}\\
        
        \includegraphics[width=0.8\linewidth]{images/examples/cmap_ssr.png} & 
        \includegraphics[width=0.8\linewidth]{images/examples/DL_cmap.png}\\
        
        Prediction from Koch's filters & Prediction from U-Net model \\
        \includegraphics[width=0.8\linewidth]{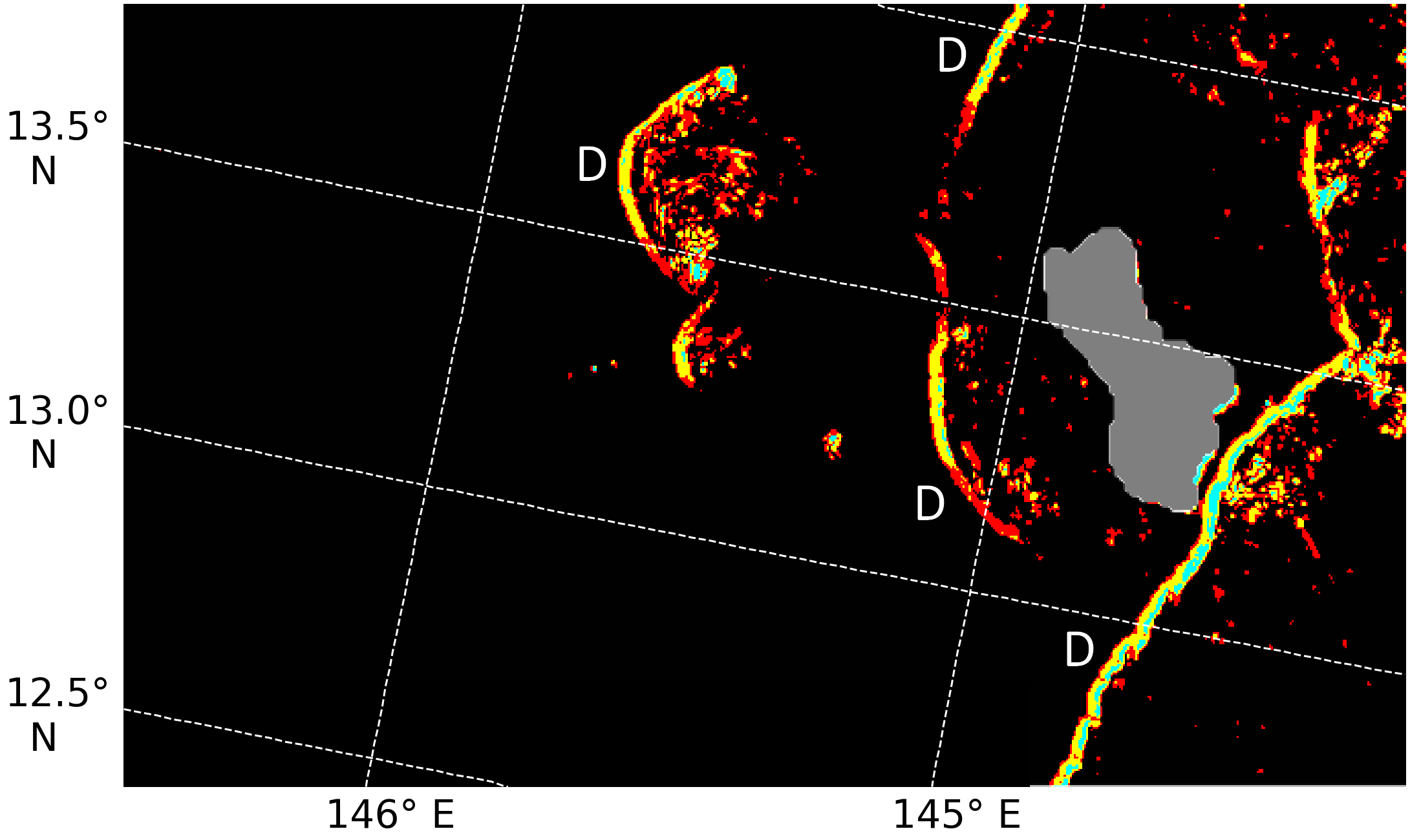} & 
        \includegraphics[width=0.8\linewidth]{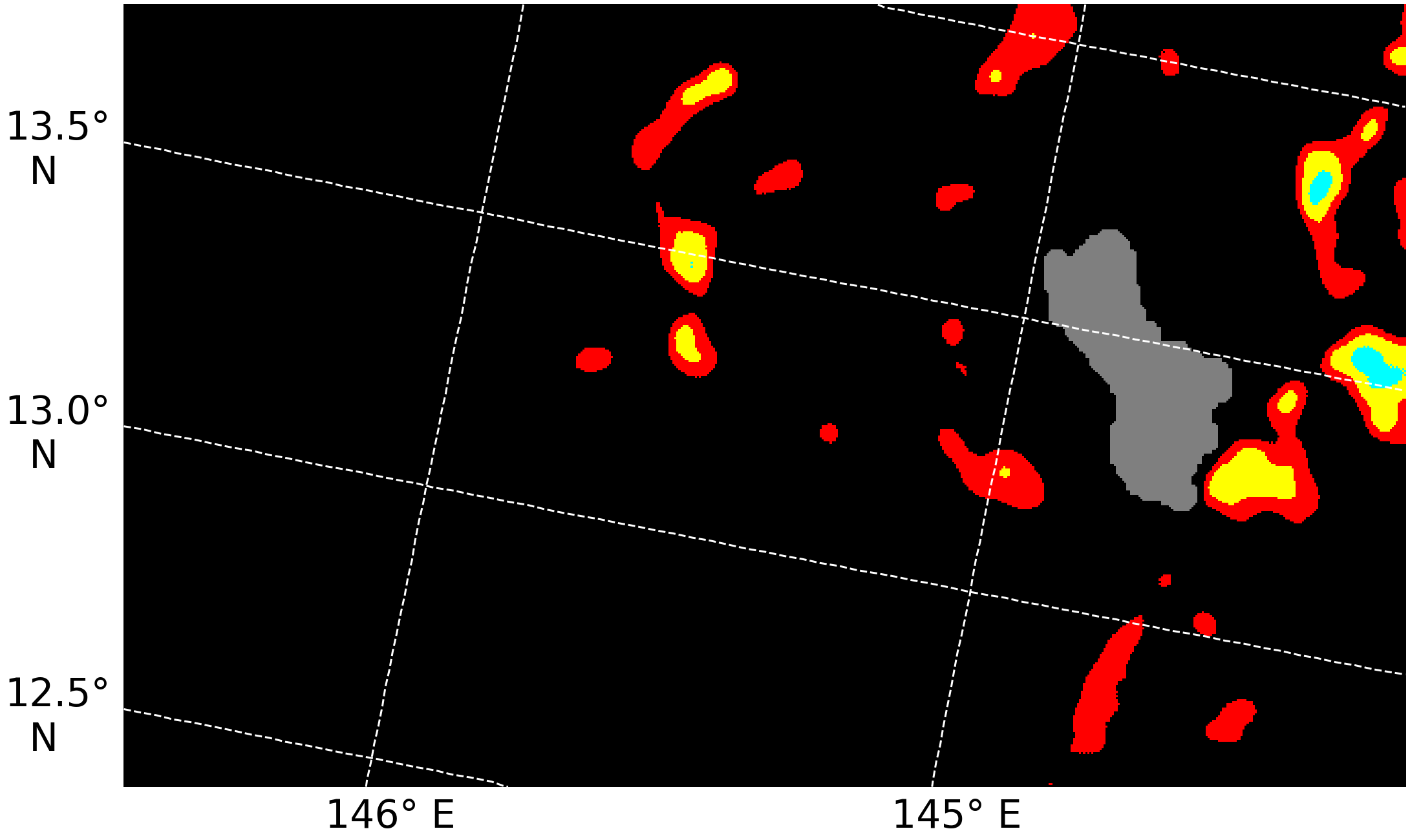}\\
    \end{tabularx}}
    \caption{\textcolor{black}{Example of SAR-derived reflectivity estimation from the observation from August 05\textsuperscript{th} 2018, at 20:07:39.}}
    \label{fig:NEXRAD_examples3}
\end{figure*}

\begin{figure*}[!ht]
    \small
    \centering
    \resizebox{\linewidth}{!}{%
    \begin{tabularx}{\linewidth}{YY}
        SAR observation & NEXRAD reflectivity (groundtruth)  \\
        \includegraphics[width=0.8\linewidth]{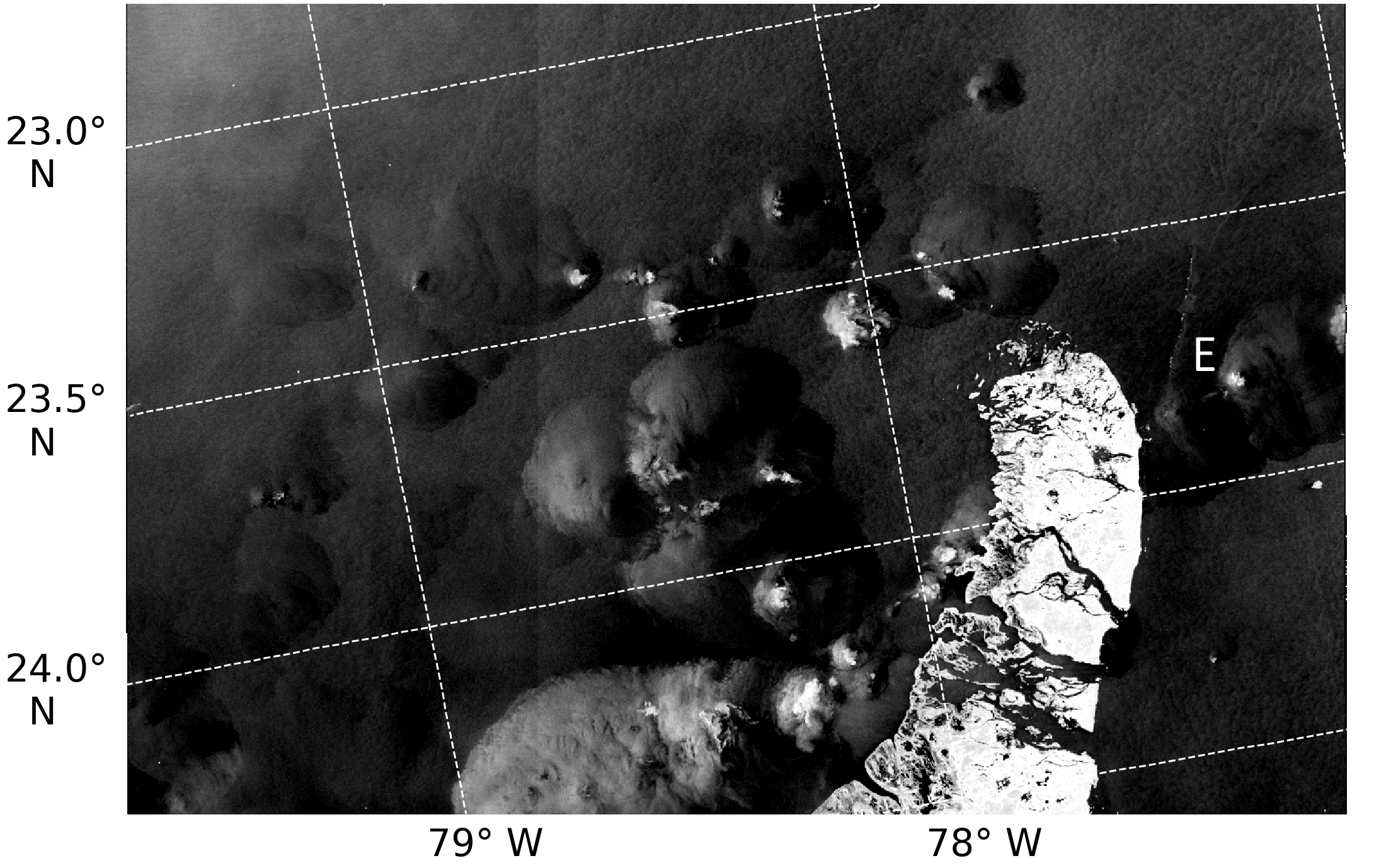} & 
        \includegraphics[width=0.8\linewidth]{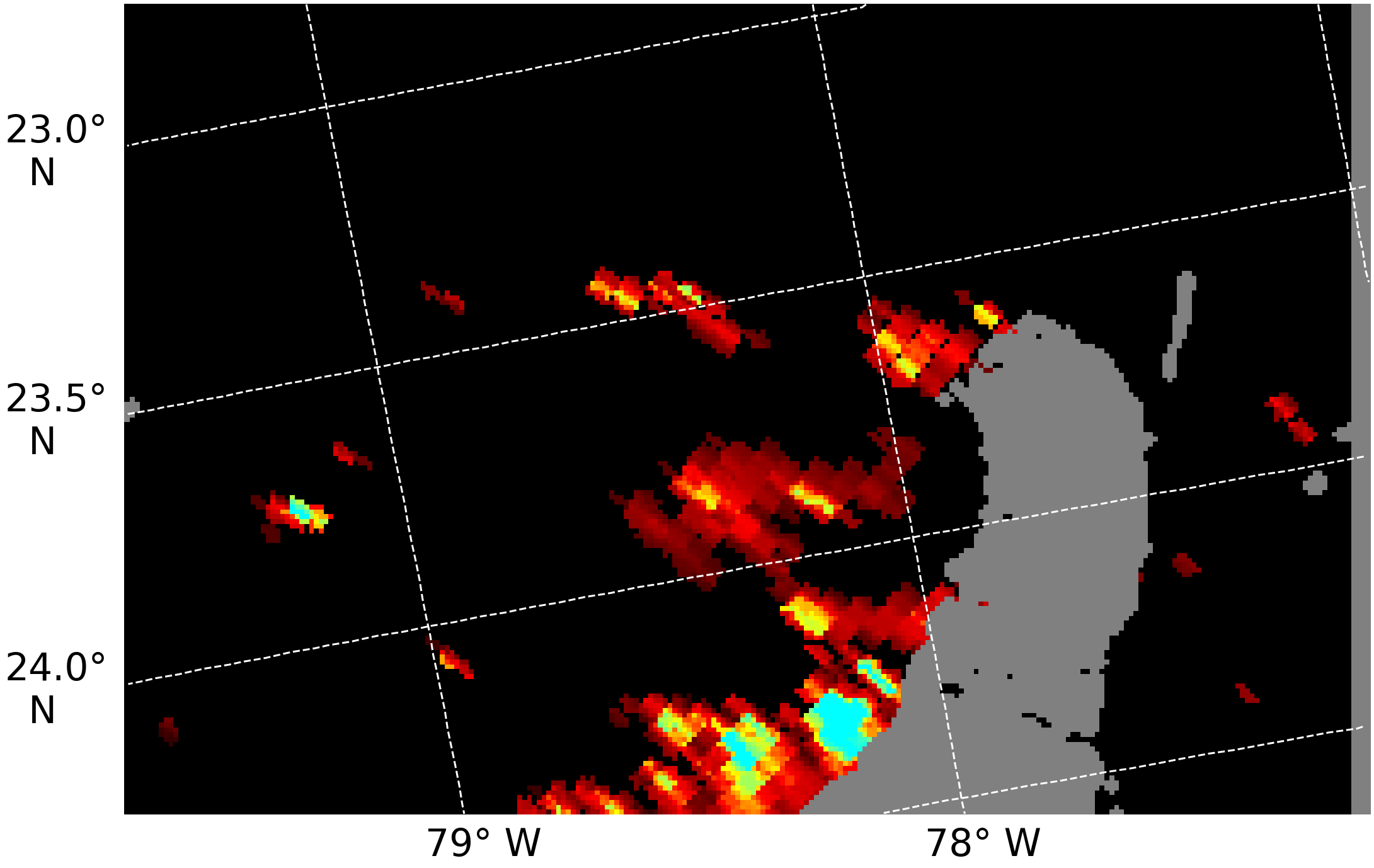}\\
        
        \includegraphics[width=0.8\linewidth]{images/examples/cmap_ssr.png} & 
        \includegraphics[width=0.8\linewidth]{images/examples/DL_cmap.png}\\
        
        Prediction from Koch's filters & Prediction from U-Net model\\
        \includegraphics[width=0.8\linewidth]{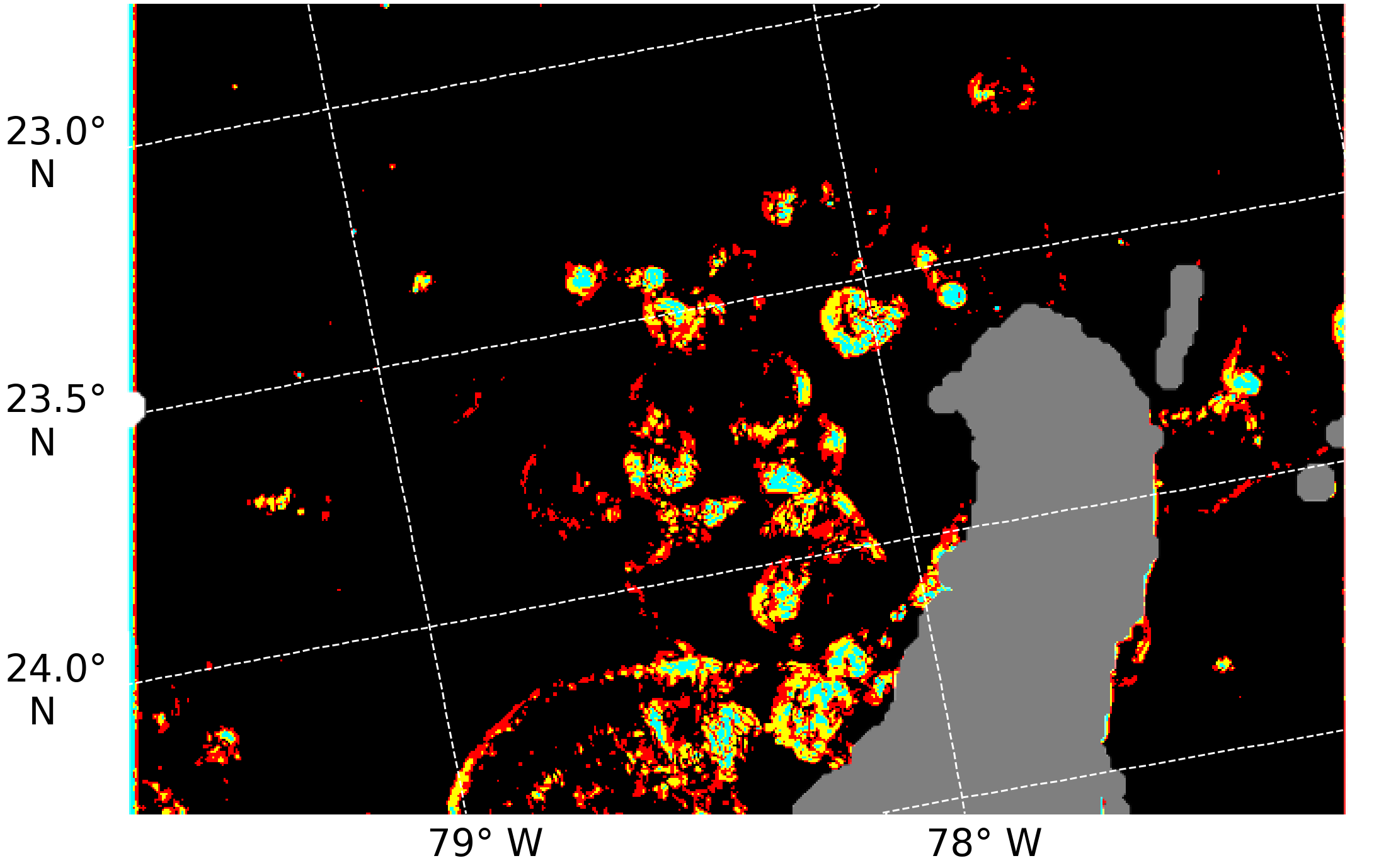} & 
        \includegraphics[width=0.8\linewidth]{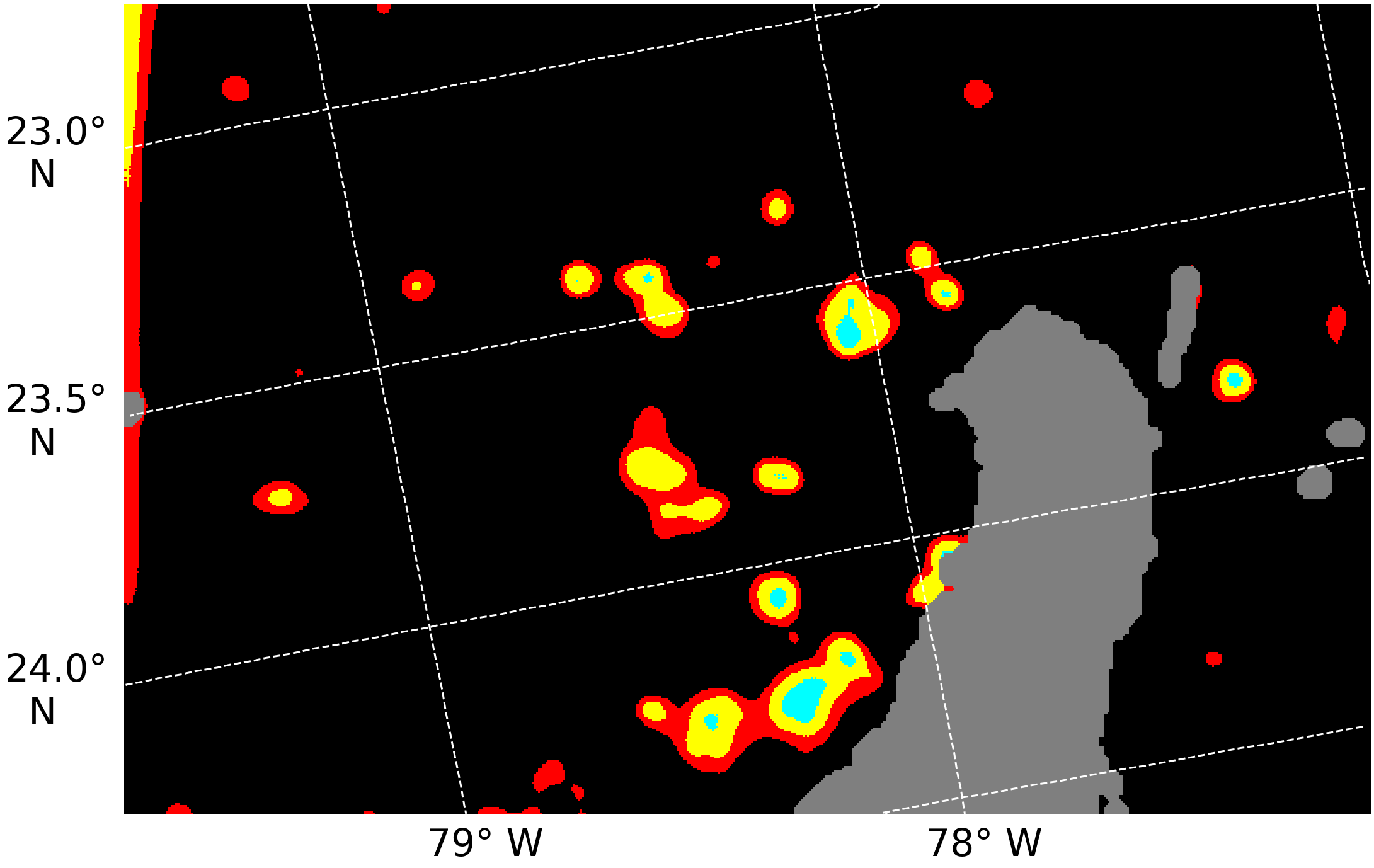}\\
    \end{tabularx}}
    \caption{\textcolor{black}{Example of SAR-derived reflectivity estimation from the observation from August 19\textsuperscript{th} 2018 at 23:19:09.}}
    \label{fig:NEXRAD_examples2}
\end{figure*}

Figure\textcolor{black}{s \ref{fig:NEXRAD_examples1} to \ref{fig:NEXRAD_examples3}} shows some examples of rainfall predictions using SAR data and either the U-Net architecture or the fine-tuned Koch's filters. Overall, the SAR rainfall predictions appear to concord with the NEXRAD acquisitions over the ocean, with different sensitivities.
In Figure \textcolor{black}{\ref{fig:NEXRAD_examples1}}, the fine-tuned filter well detects the rainy regions but indicate less or no rain \textcolor{black}{in the main rain cell higher than 25° N, indicated by the letter 'A'}. This is mainly due by the direct use of high-pass filters while the U-Net architecture is more general. Also, the fine-tuned filter tends to detects smaller rain patches, not detected in NEXRAD. Three neighbouring dots located \textcolor{black}{near the letter 'B' are false positives and} correspond to 3 ships.
\textcolor{black}{In the top-left corner, near the letter 'C', the weather radar observes reflectivities lower than the 24.7 dBZ threshold, which correlates with low temperature brightness in the 11.2 µm channel from GOES-16's radiometer, hinting to the presence of a deep cloud cover. This indicates that low rain rates are not visible on the SAR observation, either because they evaporate in the air column or because their impact on the ocean surface is too weak.}

In Figure  \textcolor{black}{\ref{fig:NEXRAD_examples3}}, the fine-tuned filter wrongly interprets gust fronts \textcolor{black}{(letters 'D')} as strong rain due to their strong discontinuity with respect to the background radar signal.
In Figure  \textcolor{black}{\ref{fig:NEXRAD_examples2}}, we illustrate limitations of the NEXRAD system as it is unable to detect the rain patches located on the right-hand side \textcolor{black}{(letter 'E')}, possibly due to masking by the topography. \textcolor{black}{The observations are samples of the test set.}

\begin{figure*}[!h]
    \small
    \centering
    \resizebox{\linewidth}{!}{%
    \begin{tabularx}{\linewidth}{kYY}
    \footnotesize
         & 2018-09-29 09:28:26 & 2021-10-16 21:05:33 \\
 
        \rotatebox[origin=t]{90}{ \parbox{1.8cm}{\centering SAR\\observation }} &
        \includegraphics[height=0.16\paperheight]{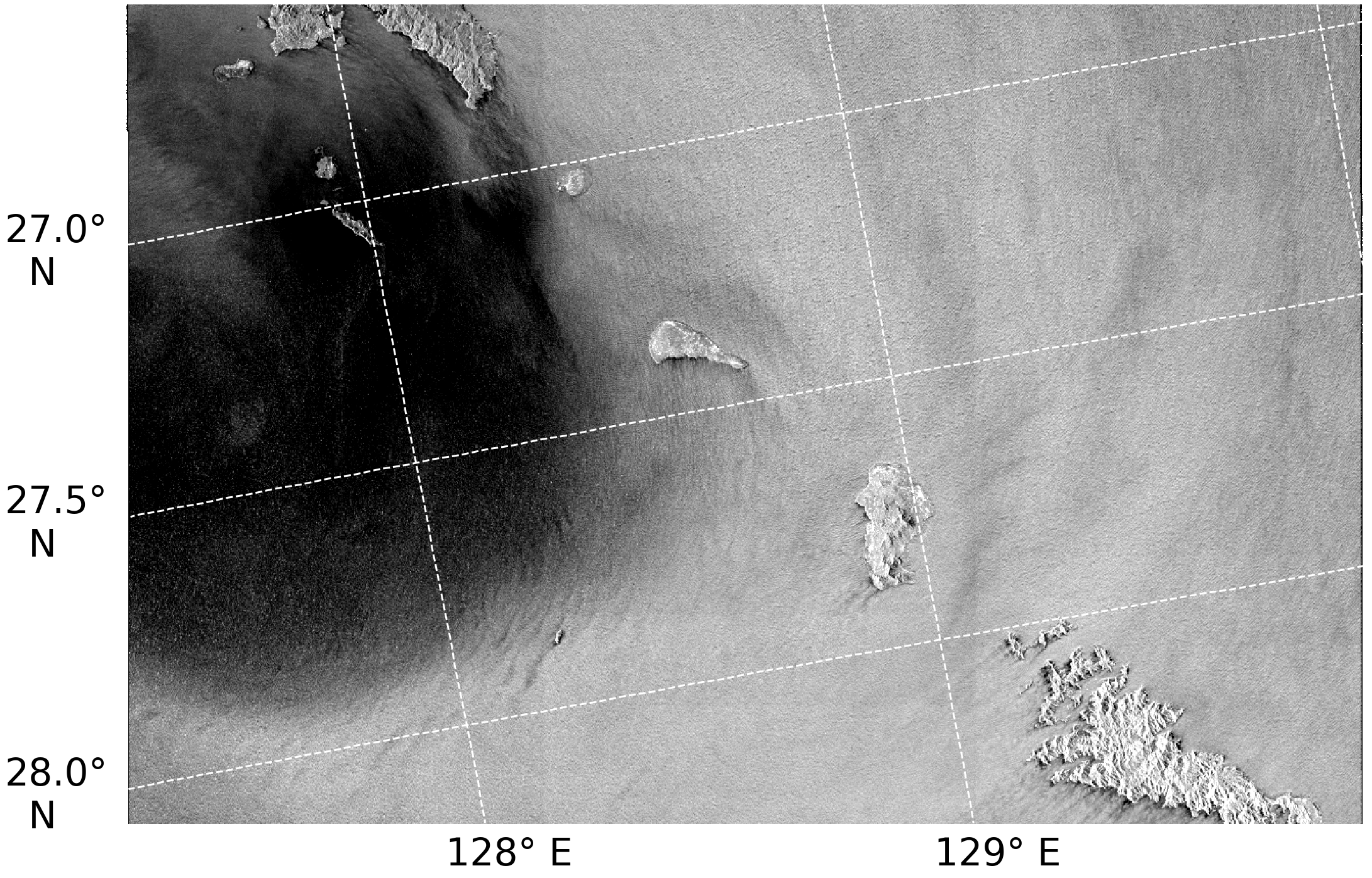} &
        \includegraphics[height=0.16\paperheight]{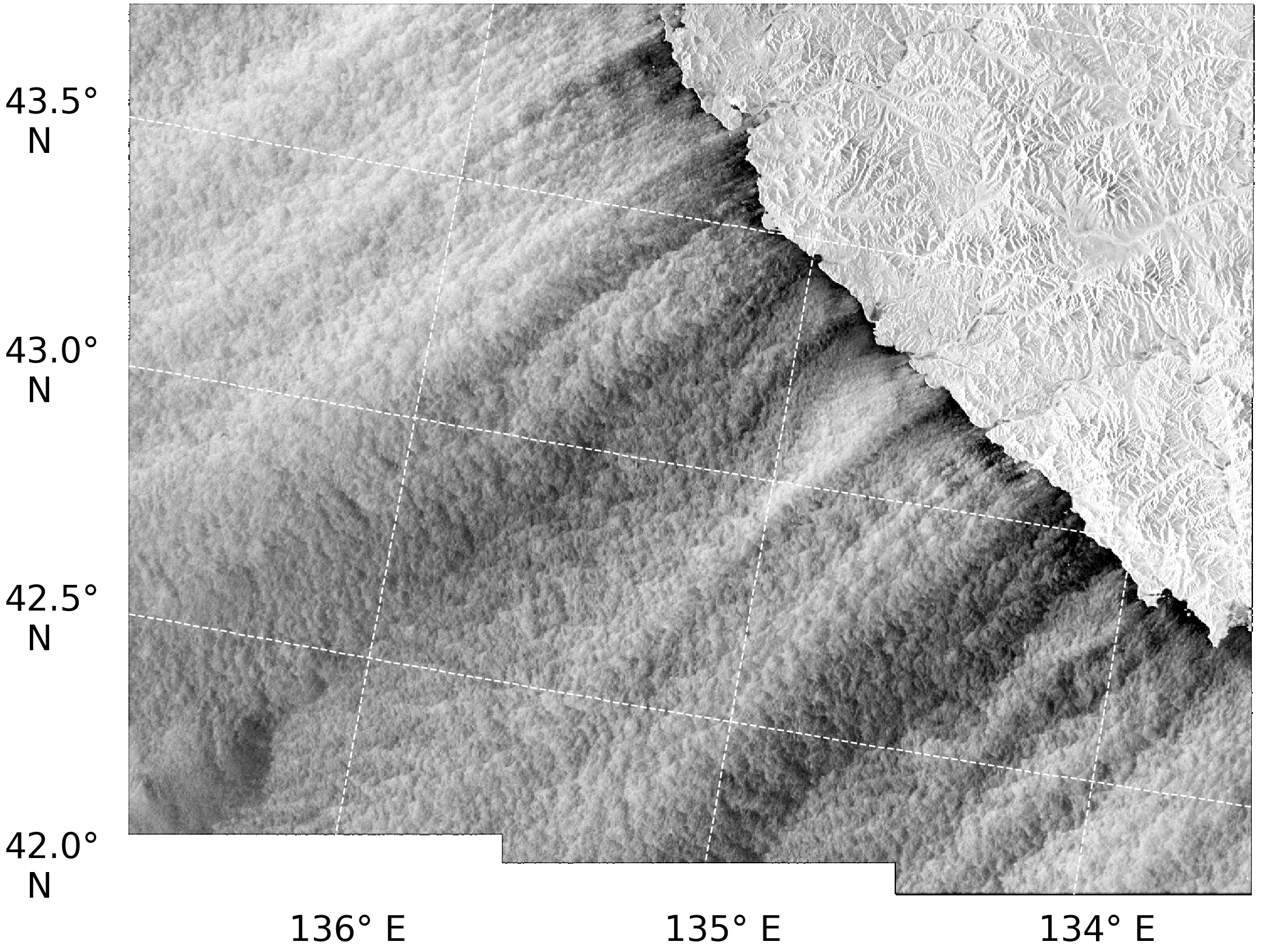} \\

        & \includegraphics[width=0.8\linewidth]{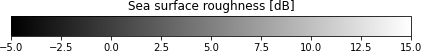}& \includegraphics[width=0.8\linewidth]{images/examples/cmap_ssr.png} \\
        
        \rotatebox[origin=t]{90}{ \parbox{1.8cm}{\centering Prediction from U-Net model.}} &
        \includegraphics[height=0.16\paperheight]{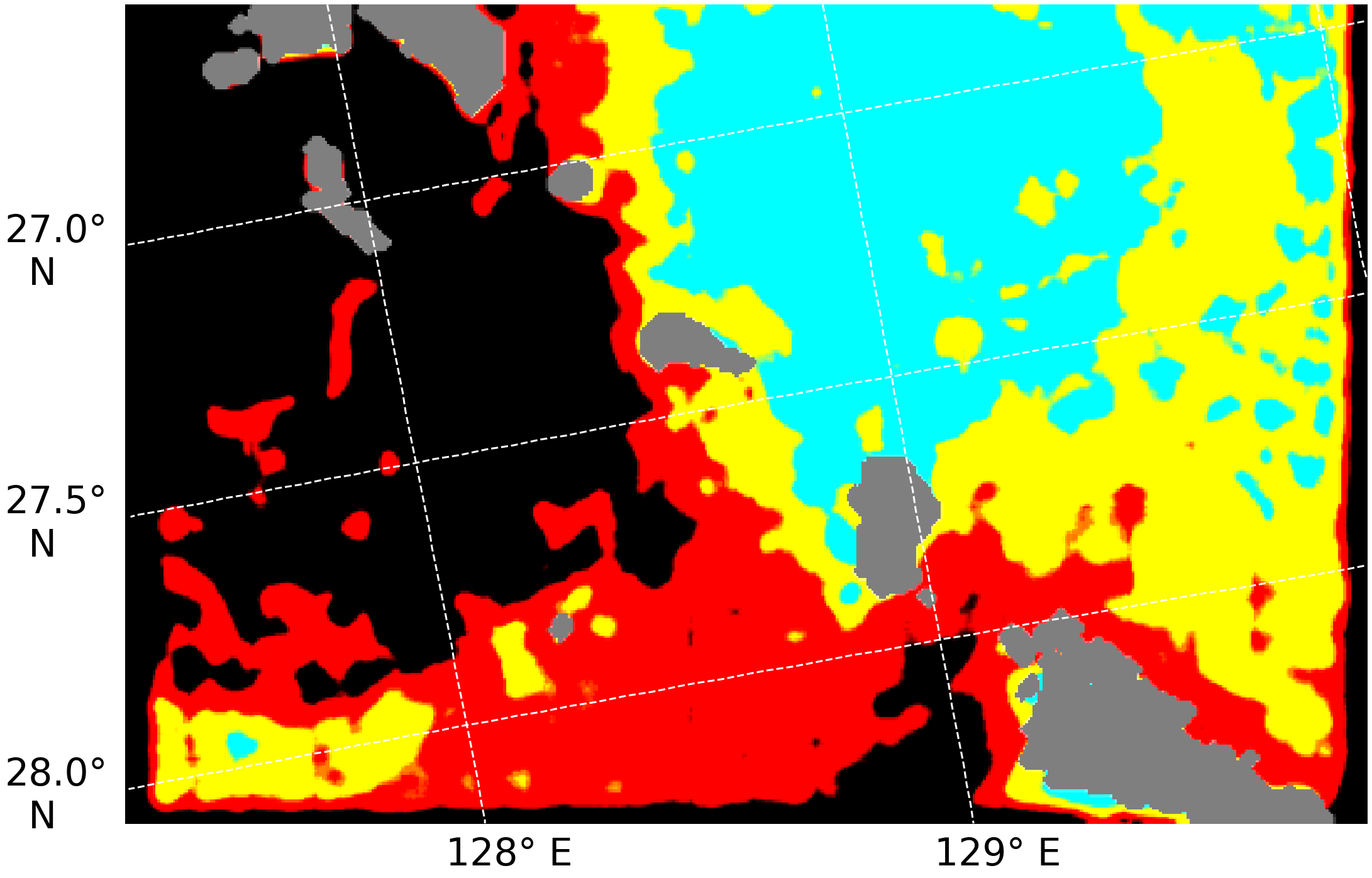} &
        \includegraphics[height=0.16\paperheight]{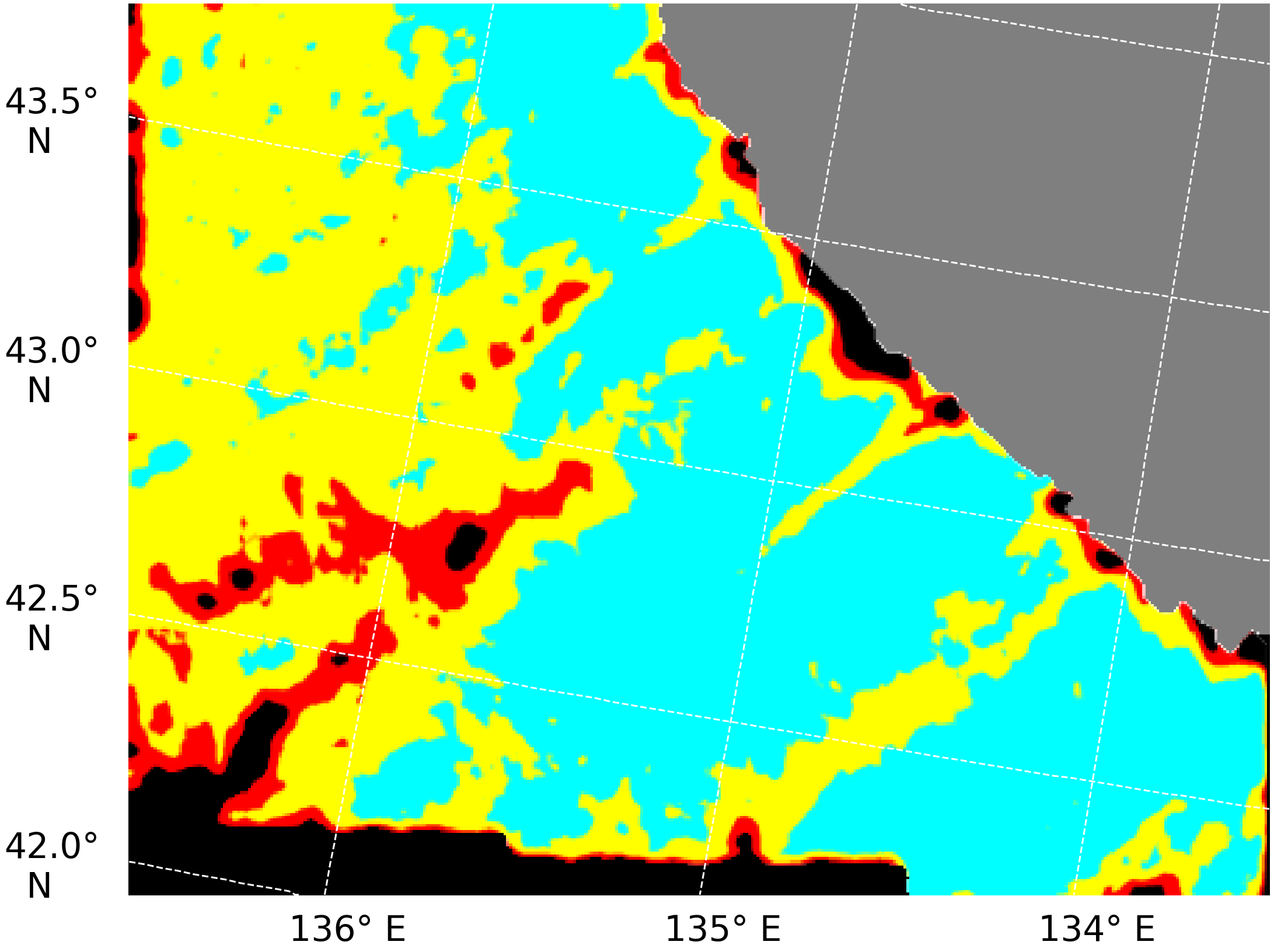} \\

        & \multicolumn{2}{c}{\includegraphics[width=0.4\linewidth]{images/examples/DL_cmap.png}} \\
        
        \rotatebox[origin=t]{90}{ \parbox{1.8cm}{\centering \textcolor{black}{SAR-based}\\wind speed }} &
        \includegraphics[height=0.16\paperheight]{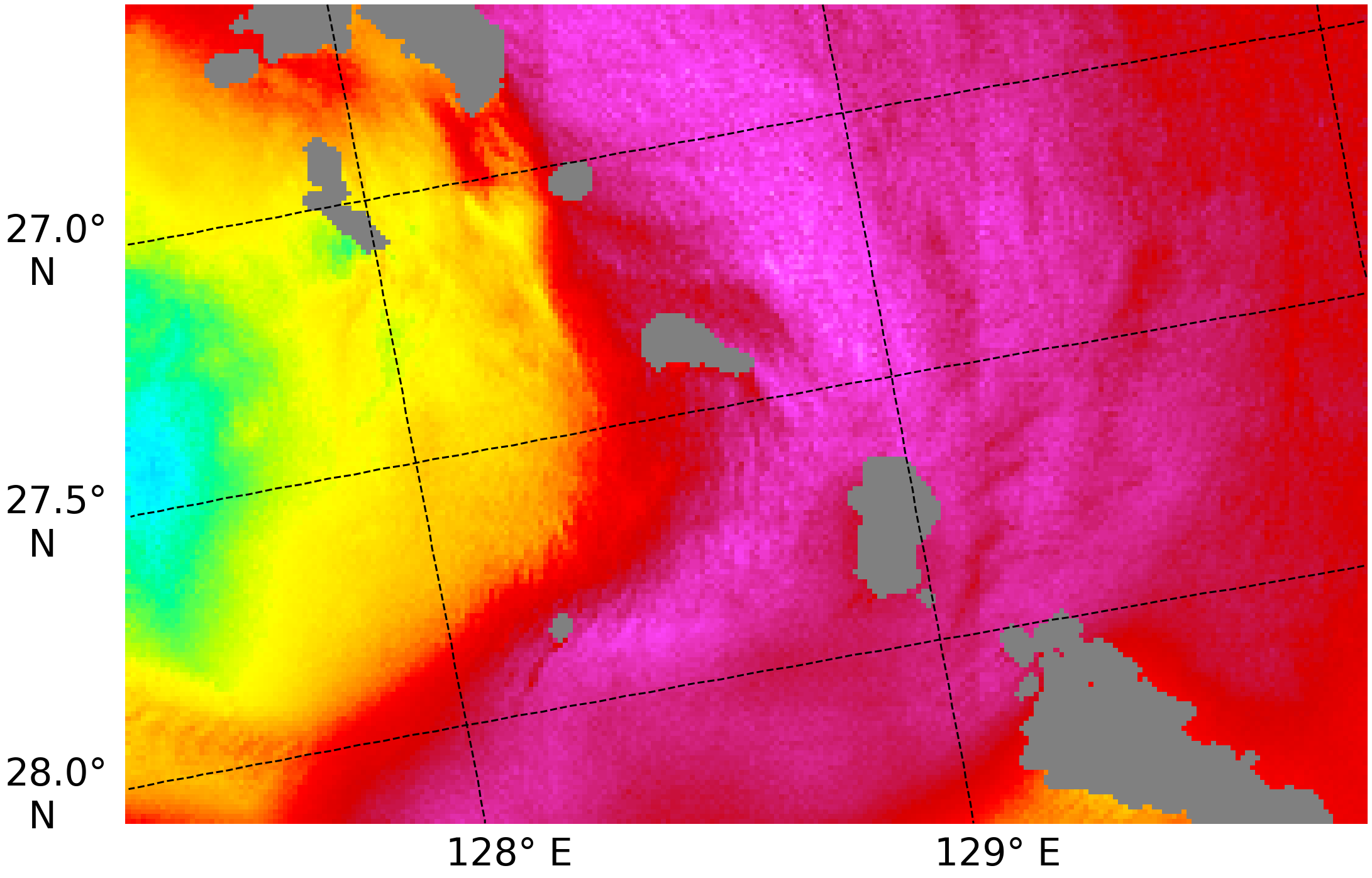} &
        \includegraphics[height=0.16\paperheight]{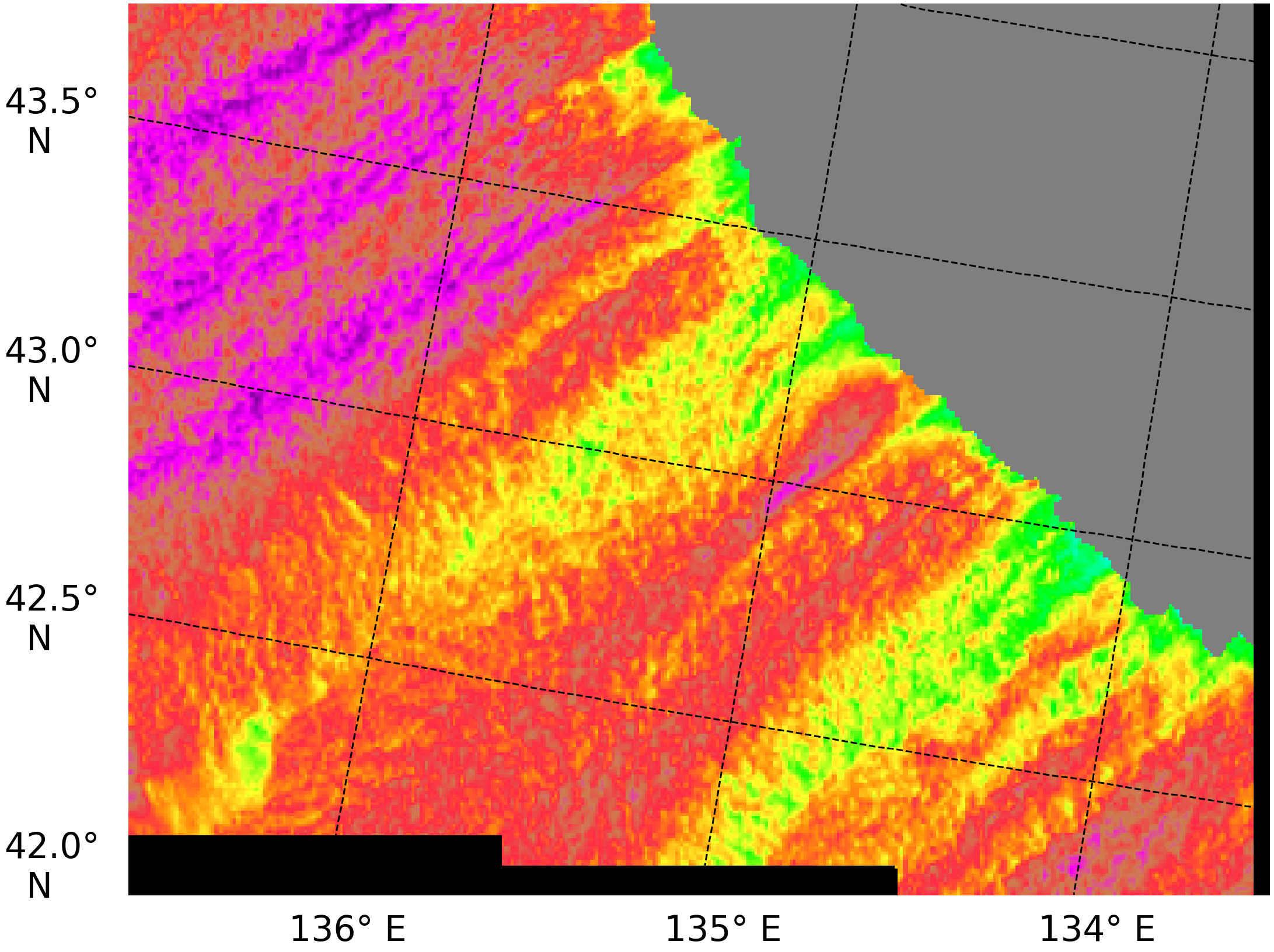} \\

        & \includegraphics[width=0.8\linewidth]{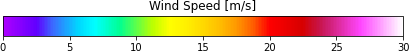} & \includegraphics[width=0.8\linewidth]{images/examples/ecmwf_cmp.png} \\

    \end{tabularx}}
    
    \caption{Examples of wrong rainfall estimation. The overestimation from September 29\textsuperscript{th} 2018 at 09:28:26 \textcolor{black}{(left)} is linked to the strong wind of Typhon Trami. The overestimation from October 16\textsuperscript{th} 2021 at 21:05:33 \textcolor{black}{(right)} is caused by topography-induced wind regimes. Lands are greyed out.}
    \label{fig:cherry_picked_bad_examples}
\end{figure*}

As the dataset contains limited data for high wind speed, the model is unable to estimate the rainfall on strong winds. In particular, running the model on miscellaneous IW acquisitions shows a tendency to overestimate the reflectivity. Figure \ref{fig:cherry_picked_bad_examples} indicates two cases of these overestimates. The observation from September 29\textsuperscript{th} 2018 contains wind speed around 30 m/s, higher than any data contained in the dataset. The SAR observation therefore appear particularly bright and led the rainfall estimation to reach the 38.8 dBZ threshold on a large area. The observation from October 16\textsuperscript{th} 2021 depicts strong mountain winds around 12 m/s. The topography of the coast generates are strong gradient parallel to the coast, whereas gravity waves cause the SAR intensity to vary parallel to the wind direction. The cumulative signatures are recognized as strong rainfall by the model. No weather radar groundtruth was available for these observations, but the distinctive signature of precipitation do not appear on the SAR observation.  


\section{Conclusion}

The monitoring of rain over the oceans is a key challenge for weather modeling and forecasting. This is particularly important for flood mitigation in coastal areas. While land-based sensors cannot monitor the open ocean, the satellite-derived retrieval of rain rate remains a challenge, especially at high resolution, despite the variety of rain-impacted and rain-measuring spaceborne instruments. In this respect, the effect of precipitation on satellite SAR observations of the sea surface has been widely documented.

This study demonstrates that deep learning opens new avenues for the \textcolor{black}{monitoring} of sea surface rain \textcolor{black}{patterns} at high resolution from satellite SAR observations. We exploit a state-of-the-art image-to-image translation architecture, namely a U-Net. The training scheme relies on a collocated dataset of NEXRAD weather radar data and Sentinel-1 SAR observations. The deep learning model is compared to a neural network implementation of the Koch's filters. We report an accurate segmentation of rainy areas at sea surface and satisfactory ability to discriminate rain between 24.7 dBZ and 31.5 dBZ and 38.8 dBZ and above 38.8 dBZ. The proposed approach outperforms previous work based on Koch's filters and points out the importance of a realignment-based preprocessing of the training dataset.


Future work could therefore benefit from the generalisation of the proposed approach to other SAR modes such as WV modes which may involve other incidence angle ranges. The addition of other SAR parameters, such as the VH-polarization, could increase the performances, especially at high wind speed where VV-polarization is known to saturate. However, an extension to high wind speed would need new collocations dedicated to these wind regimes. Finally, the study also supports the development of a joint wind speed and rain rate retrieval at sea surface. Both are currently done independently despite each phenomena impacting the aspect of the SAR signature of the other.


\section*{Acknowledgment}

We thank Alexis Mouche (Laboratoire d’Océanographie Physique et Spatiale, Ifremer) for the access to collocations between NEXRAD and Sentinel-1.

\clearpage

\bibliographystyle{IEEEtran}
\bibliography{main}

\begin{thebibliography}{10}
\providecommand{\url}[1]{#1}
\csname url@samestyle\endcsname
\providecommand{\newblock}{\relax}
\providecommand{\bibinfo}[2]{#2}
\providecommand{\BIBentrySTDinterwordspacing}{\spaceskip=0pt\relax}
\providecommand{\BIBentryALTinterwordstretchfactor}{4}
\providecommand{\BIBentryALTinterwordspacing}{\spaceskip=\fontdimen2\font plus
\BIBentryALTinterwordstretchfactor\fontdimen3\font minus
  \fontdimen4\font\relax}
\providecommand{\BIBforeignlanguage}[2]{{%
\expandafter\ifx\csname l@#1\endcsname\relax
\typeout{** WARNING: IEEEtran.bst: No hyphenation pattern has been}%
\typeout{** loaded for the language `#1'. Using the pattern for}%
\typeout{** the default language instead.}%
\else
\language=\csname l@#1\endcsname
\fi
#2}}
\providecommand{\BIBdecl}{\relax}
\BIBdecl

\bibitem{10.1029/2019ms001618}
\BIBentryALTinterwordspacing
X.~Li, J.~R. Mecikalski, J.~Srikishen, B.~Zavodsky, and W.~A. Petersen,
  ``Assimilation of {GPM} rain rate products with {GSI} data assimilation
  system for heavy and light precipitation events,'' \emph{Journal of Advances
  in Modeling Earth Systems}, vol.~12, no.~5, May 2020. [Online]. Available:
  \url{https://doi.org/10.1029/2019ms001618}
\BIBentrySTDinterwordspacing

\bibitem{douville2021}
H.~Douville, K.~Raghavan, J.~Renwick, R.~Allan, P.~Arias, M.~Barlow,
  R.~Cerezo-Mota, A.~Cherchi, T.~Gan, J.~Gergis, D.~Jiang, A.~Khan, W.~P. Mba,
  D.~Rosenfeld, J.~Tierney, and O.~Zolina, ``2021 : Water cycle changes,'' in
  \emph{Climate Change 2021: The Physical Science Basis. Contribution of
  Working Group I to the Sixth Assessment Report of the Intergovernmental Panel
  on Climate Change}.\hskip 1em plus 0.5em minus 0.4em\relax Cambridge
  University Press, 2021.

\bibitem{10.3390/rs5115702}
\BIBentryALTinterwordspacing
E.~de~Coning, ``Optimizing satellite-based precipitation estimation for
  nowcasting of rainfall and flash flood events over the south african
  domain,'' \emph{Remote Sensing}, vol.~5, no.~11, pp. 5702--5724, Nov. 2013.
  [Online]. Available: \url{https://doi.org/10.3390/rs5115702}
\BIBentrySTDinterwordspacing

\bibitem{10.1109/tgrs.2007.903685}
\BIBentryALTinterwordspacing
F.~S. Marzano, G.~Rivolta, E.~Coppola, B.~Tomassetti, and M.~Verdecchia,
  ``Rainfall nowcasting from multisatellite passive-sensor images using a
  recurrent neural network,'' \emph{{IEEE} Transactions on Geoscience and
  Remote Sensing}, vol.~45, no.~11, pp. 3800--3812, Nov. 2007. [Online].
  Available: \url{https://doi.org/10.1109/tgrs.2007.903685}
\BIBentrySTDinterwordspacing

\bibitem{10.1016/j.rse.2015.02.006}
\BIBentryALTinterwordspacing
K.~Topouzelis and D.~Kitsiou, ``Detection and classification of mesoscale
  atmospheric phenomena above sea in {SAR} imagery,'' \emph{Remote Sensing of
  Environment}, vol. 160, pp. 263--272, Apr. 2015. [Online]. Available:
  \url{https://doi.org/10.1016/j.rse.2015.02.006}
\BIBentrySTDinterwordspacing

\bibitem{ayet2021uncovering}
A.~Ayet, N.~Rascle, B.~Chapron, F.~Couvreux, and L.~Terray, ``Uncovering
  air-sea interaction in oceanic submesoscale frontal regions using
  high-resolution satellite observations,'' \emph{US Clivar variations},
  vol.~19, no.~1, 2021.

\bibitem{10.1175/1520-0469}
\BIBentryALTinterwordspacing
J.~S. Marshall, R.~C. Langille, and W.~M.~K. Palmer, ``{MEASUREMENT} {OF}
  {RAINFALL} {BY} {RADAR},'' \emph{Journal of Meteorology}, vol.~4, no.~6, pp.
  186--192, Dec. 1947. [Online]. Available:
  \url{https://doi.org/10.1175/1520-0469(1947)004<0186:morbr>2.0.co;2}
\BIBentrySTDinterwordspacing

\bibitem{10.1175/jhm-d-19-0194.1}
\BIBentryALTinterwordspacing
J.~Zhang, L.~Tang, S.~Cocks, P.~Zhang, A.~Ryzhkov, K.~Howard, C.~Langston, and
  B.~Kaney, ``A dual-polarization radar synthetic {QPE} for operations,''
  \emph{Journal of Hydrometeorology}, vol.~21, no.~11, pp. 2507--2521, Nov.
  2020. [Online]. Available: \url{https://doi.org/10.1175/jhm-d-19-0194.1}
\BIBentrySTDinterwordspacing

\bibitem{10.1016/b0-12-227090-8}
\BIBentryALTinterwordspacing
G.~Liu, ``{SATELLITE} {REMOTE} {SENSING} $\vert$ precipitation,'' in
  \emph{Encyclopedia of Atmospheric Sciences}.\hskip 1em plus 0.5em minus
  0.4em\relax Elsevier, 2003, pp. 1972--1979. [Online]. Available:
  \url{https://doi.org/10.1016/b0-12-227090-8/00352-3}
\BIBentrySTDinterwordspacing

\bibitem{10.2151/jmsj.2021-011}
\BIBentryALTinterwordspacing
S.~SETO, T.~IGUCHI, R.~MENEGHINI, J.~AWAKA, T.~KUBOTA, T.~MASAKI, and
  N.~TAKAHASHI, ``The precipitation rate retrieval algorithms for the {GPM}
  dual-frequency precipitation radar,'' \emph{Journal of the Meteorological
  Society of Japan. Ser. {II}}, vol.~99, no.~2, pp. 205--237, 2021. [Online].
  Available: \url{https://doi.org/10.2151/jmsj.2021-011}
\BIBentrySTDinterwordspacing

\bibitem{10.1007/978-3-030-24568-9_19}
\BIBentryALTinterwordspacing
G.~J. Huffman, D.~T. Bolvin, D.~Braithwaite, K.-L. Hsu, R.~J. Joyce, C.~Kidd,
  E.~J. Nelkin, S.~Sorooshian, E.~F. Stocker, J.~Tan, D.~B. Wolff, and P.~Xie,
  ``Integrated multi-satellite retrievals for the global precipitation
  measurement ({GPM}) mission ({IMERG}),'' in \emph{Advances in Global Change
  Research}.\hskip 1em plus 0.5em minus 0.4em\relax Springer International
  Publishing, 2020, pp. 343--353. [Online]. Available:
  \url{https://doi.org/10.1007/978-3-030-24568-9_19}
\BIBentrySTDinterwordspacing

\bibitem{10.1175/1525-7541(2004)005<0487:CAMTPG>2.0.CO;2}
\BIBentryALTinterwordspacing
R.~J. Joyce, J.~E. Janowiak, P.~A. Arkin, and P.~Xie, ``{CMORPH}: A method that
  produces global precipitation estimates from passive microwave and infrared
  data at high spatial and temporal resolution,'' \emph{Journal of
  Hydrometeorology}, vol.~5, no.~3, pp. 487--503, Jun. 2004. [Online].
  Available:
  \url{https://doi.org/10.1175/1525-7541(2004)005<0487:camtpg>2.0.co;2}
\BIBentrySTDinterwordspacing

\bibitem{jackson2004synthetic}
C.~Jackson, \emph{Synthetic aperture radar : marine user's manual}.\hskip 1em
  plus 0.5em minus 0.4em\relax Washington, D.C: U.S. Dept. of Commerce,
  National Oceanic and Atmospheric Administration, National Environmental
  Satellite, Data, and Information Serve, Office of Research and Applications,
  2004.

\bibitem{10.1109/IGARSS.1996.516666}
C.~Melshelmer, W.~Alpers, and M.~Gade, ``Investigation of
  multifrequency/multipolarization radar signatures of rain cells, derived from
  sir-c/x-sar data,'' in \emph{IGARSS '96. 1996 International Geoscience and
  Remote Sensing Symposium}, vol.~2, 1996, pp. 1370--1372 vol.2.

\bibitem{10.1117/12.373044}
\BIBentryALTinterwordspacing
P.~Clemente-Colon, P.~C. Manousos, W.~G. Pichel, and K.~S. Friedman,
  ``{Observations of Hurricane Bonnie in spaceborne synthetic aperture radar
  (SAR) and next-generation Doppler weather radar (NEXRAD)},'' in
  \emph{Satellite Remote Sensing of Clouds and the Atmosphere IV}, J.~E.
  Russell, Ed., vol. 3867, International Society for Optics and
  Photonics.\hskip 1em plus 0.5em minus 0.4em\relax SPIE, 1999, pp. 63 -- 70.
  [Online]. Available: \url{https://doi.org/10.1117/12.373044}
\BIBentrySTDinterwordspacing

\bibitem{10.1109/36.921411}
I.-I. Lin, W.~Alpers, V.~Khoo, H.~Lim, T.~Lim, and D.Kasilingam, ``An ers-1
  synthetic aperture radar image of a tropical squall line compared with
  weather radar data,'' \emph{IEEE Transactions on Geoscience and Remote
  Sensing}, vol.~39, no.~5, pp. 937--945, 2001.

\bibitem{10.1029/2000JC000263}
C.~Melsheimer, W.~Alpers, and M.~Gade, ``Simultaneous observations of rain
  cells over the ocean by the synthetic aperture radar aboard the ers
  satellites and by surface-based weather radars,'' \emph{Journal of
  Geophysical Research}, vol. 106, pp. 4665--4678, 03 2001.

\bibitem{Portabella2001}
\BIBentryALTinterwordspacing
M.~Portabella and A.~Stoffelen, ``Rain detection and quality control of
  {SeaWinds},'' \emph{Journal of Atmospheric and Oceanic Technology}, vol.~18,
  no.~7, pp. 1171--1183, Jul. 2001. [Online]. Available:
  \url{https://doi.org/10.1175/1520-0426(2001)018<1171:rdaqco>2.0.co;2}
\BIBentrySTDinterwordspacing

\bibitem{10.1029/JC095iC10p18353}
\BIBentryALTinterwordspacing
J.~A. Nystuen, ``A note on the attenuation of surface gravity waves by
  rainfall,'' \emph{Journal of Geophysical Research}, vol.~95, no. C10, p.
  18353, 1990. [Online]. Available:
  \url{https://doi.org/10.1029/jc095ic10p18353}
\BIBentrySTDinterwordspacing

\bibitem{10.1109/igarss.2008.4778801}
\BIBentryALTinterwordspacing
E.~Attema, P.~Snoeij, M.~Davidson, N.~Floury, G.~Levrini, B.~Rommen, and
  B.~Rosich, ``The european {GMES} sentinel-1 radar mission,'' in
  \emph{{IGARSS} 2008 - 2008 {IEEE} International Geoscience and Remote Sensing
  Symposium}.\hskip 1em plus 0.5em minus 0.4em\relax {IEEE}, 2008. [Online].
  Available: \url{https://doi.org/10.1109/igarss.2008.4778801}
\BIBentrySTDinterwordspacing

\bibitem{10.1109/tgrs.2008.2001032}
\BIBentryALTinterwordspacing
D.~E. Weissman and M.~A. Bourassa, ``Measurements of the effect of rain-induced
  sea surface roughness on the {QuikSCAT} scatterometer radar cross section,''
  \emph{{IEEE} Transactions on Geoscience and Remote Sensing}, vol.~46, no.~10,
  pp. 2882--2894, Oct. 2008. [Online]. Available:
  \url{https://doi.org/10.1109/tgrs.2008.2001032}
\BIBentrySTDinterwordspacing

\bibitem{10.1109/tgrs.2019.2953143}
\BIBentryALTinterwordspacing
X.~Chen, W.~Huang, C.~Zhao, and Y.~Tian, ``Rain detection from x-band marine
  radar images: A support vector machine-based approach,'' \emph{{IEEE}
  Transactions on Geoscience and Remote Sensing}, vol.~58, no.~3, pp.
  2115--2123, Mar. 2020. [Online]. Available:
  \url{https://doi.org/10.1109/tgrs.2019.2953143}
\BIBentrySTDinterwordspacing

\bibitem{10.1109/tgrs.2014.2367654}
\BIBentryALTinterwordspacing
F.~Xu, X.~Li, P.~Wang, J.~Yang, W.~G. Pichel, and Y.-Q. Jin, ``A backscattering
  model of rainfall over rough sea surface for synthetic aperture radar,''
  \emph{{IEEE} Transactions on Geoscience and Remote Sensing}, vol.~53, no.~6,
  pp. 3042--3054, Jun. 2015. [Online]. Available:
  \url{https://doi.org/10.1109/tgrs.2014.2367654}
\BIBentrySTDinterwordspacing

\bibitem{10.1002/gdj3.73}
\BIBentryALTinterwordspacing
C.~Wang, A.~Mouche, P.~Tandeo, J.~E. Stopa, N.~Long{\'{e}}p{\'{e}}, G.~Erhard,
  R.~C. Foster, D.~Vandemark, and B.~Chapron, ``A labelled ocean {SAR} imagery
  dataset of ten geophysical phenomena from sentinel-1 wave mode,''
  \emph{Geoscience Data Journal}, vol.~6, no.~2, pp. 105--115, Jul. 2019.
  [Online]. Available: \url{https://doi.org/10.1002/gdj3.73}
\BIBentrySTDinterwordspacing

\bibitem{Colin2022}
\BIBentryALTinterwordspacing
A.~Colin, R.~Fablet, P.~Tandeo, R.~Husson, C.~Peureux, N.~Long{\'{e}}p{\'{e}},
  and A.~Mouche, ``Semantic segmentation of metoceanic processes using {SAR}
  observations and deep learning,'' \emph{Remote Sensing}, vol.~14, no.~4, p.
  851, Feb. 2022. [Online]. Available: \url{https://doi.org/10.3390/rs14040851}
\BIBentrySTDinterwordspacing

\bibitem{10.1016/j.rse.2016.10.015}
\BIBentryALTinterwordspacing
W.~Alpers, B.~Zhang, A.~Mouche, K.~Zeng, and P.~W. Chan, ``Rain footprints on
  c-band synthetic aperture radar images of the ocean - revisited,''
  \emph{Remote Sensing of Environment}, vol. 187, pp. 169--185, Dec. 2016.
  [Online]. Available: \url{https://doi.org/10.1016/j.rse.2016.10.015}
\BIBentrySTDinterwordspacing

\bibitem{rs13163155}
\BIBentryALTinterwordspacing
Y.~Zhao, N.~Longépé, A.~Mouche, and R.~Husson, ``Automated rain detection by
  dual-polarization sentinel-1 data,'' \emph{Remote Sensing}, vol.~13, no.~16,
  2021. [Online]. Available: \url{https://www.mdpi.com/2072-4292/13/16/3155}
\BIBentrySTDinterwordspacing

\bibitem{10.3390/rs10121929}
\BIBentryALTinterwordspacing
X.-M. Li, T.~Zhang, B.~Huang, and T.~Jia, ``Capabilities of chinese gaofen-3
  synthetic aperture radar in selected topics for coastal and ocean
  observations,'' \emph{Remote Sensing}, vol.~10, no.~12, p. 1929, Nov. 2018.
  [Online]. Available: \url{https://doi.org/10.3390/rs10121929}
\BIBentrySTDinterwordspacing

\bibitem{10.1080/07038992.2015.1104633}
\BIBentryALTinterwordspacing
A.~A. Thompson, ``Overview of the radarsat constellation mission,''
  \emph{Canadian Journal of Remote Sensing}, vol.~41, no.~5, pp. 401--407, Sep.
  2015. [Online]. Available:
  \url{https://doi.org/10.1080/07038992.2015.1104633}
\BIBentrySTDinterwordspacing

\bibitem{10.1109/ursigass.2014.6929612}
\BIBentryALTinterwordspacing
T.~Misra and A.~S. Kirankumar, ``Risat-1: Configuration and performance
  evaluation,'' in \emph{2014 {XXXIth} {URSI} General Assembly and Scientific
  Symposium ({URSI} {GASS})}.\hskip 1em plus 0.5em minus 0.4em\relax {IEEE},
  Aug. 2014. [Online]. Available:
  \url{https://doi.org/10.1109/ursigass.2014.6929612}
\BIBentrySTDinterwordspacing

\bibitem{10.1016/j.rse.2019.111457}
\BIBentryALTinterwordspacing
C.~Wang, P.~Tandeo, A.~Mouche, J.~E. Stopa, V.~Gressani, N.~Longepe,
  D.~Vandemark, R.~C. Foster, and B.~Chapron, ``Classification of the global
  sentinel-1 {SAR} vignettes for ocean surface process studies,'' \emph{Remote
  Sensing of Environment}, vol. 234, p. 111457, Dec. 2019. [Online]. Available:
  \url{https://doi.org/10.1016/j.rse.2019.111457}
\BIBentrySTDinterwordspacing

\bibitem{owi}
\BIBentryALTinterwordspacing
A.~Mouche, P.~Vincent, and G.~Hajduch, ``Sentinel-1 ocean wind fields (owi)
  algorithm definition,'' Tech. Rep., 11 2017. [Online]. Available:
  \url{https://sentinels.copernicus.eu/documents/247904/2142675/Thermal-Denoising-of-Products-Generated-by-Sentinel-1-IPF.pdf}
\BIBentrySTDinterwordspacing

\bibitem{10.1109/TGRS.2006.873853}
F.~D. Zan and A.~M. Guarnieri, ``Topsar: Terrain observation by progressive
  scans,'' \emph{IEEE Transactions on Geoscience and Remote Sensing}, vol.~44,
  no.~9, pp. 2352--2360, 2006.

\bibitem{Sentinel1ProductSpecification}
P.~Vincent, M.~Bourbigot, H.~Johnsen, and R.~Piantanida, ``Sentinel-1 product
  specification,''
  \url{https://sentinel.esa.int/documents/247904/1877131/Sentinel-1-Product-Specification},
  2020.

\bibitem{10.1109/tgrs.2017.2765248}
\BIBentryALTinterwordspacing
J.-W. Park, A.~A. Korosov, M.~Babiker, S.~Sandven, and J.-S. Won, ``Efficient
  thermal noise removal for sentinel-1 {TOPSAR} cross-polarization channel,''
  \emph{{IEEE} Transactions on Geoscience and Remote Sensing}, vol.~56, no.~3,
  pp. 1555--1565, Mar. 2018. [Online]. Available:
  \url{https://doi.org/10.1109/tgrs.2017.2765248}
\BIBentrySTDinterwordspacing

\bibitem{10.1175/2009jtecho698.1}
\BIBentryALTinterwordspacing
H.~Hersbach, ``Comparison of c-band scatterometer {CMOD}5.n equivalent neutral
  winds with {ECMWF},'' \emph{Journal of Atmospheric and Oceanic Technology},
  vol.~27, no.~4, pp. 721--736, Apr. 2010. [Online]. Available:
  \url{https://doi.org/10.1175/2009jtecho698.1}
\BIBentrySTDinterwordspacing

\bibitem{10.1080/01431169008955114}
\BIBentryALTinterwordspacing
N.~R. DALEZIOS, ``Digital processing of weather radar signals for rainfall
  estimation,'' \emph{International Journal of Remote Sensing}, vol.~11, no.~9,
  pp. 1561--1569, Sep. 1990. [Online]. Available:
  \url{https://doi.org/10.1080/01431169008955114}
\BIBentrySTDinterwordspacing

\bibitem{NexradHandbook}
S.~Williamson, ``Doppler radar meteorological observations. part b, doppler
  radar theory and meteorology,'' FCM-H11B-2005, Office of the Federal
  Coordinator for Meteorological Services and Supporting Research, Tech. Rep.,
  2015.

\bibitem{10.1109/tgrs.2003.818811}
\BIBentryALTinterwordspacing
W.~Koch, ``Directional analysis of {SAR} images aiming at wind direction,''
  \emph{Remote Sens. Environ.}, vol.~42, no.~4, pp. 702--710, Apr. 2004.
  [Online]. Available: \url{https://doi.org/10.1109/tgrs.2003.818811}
\BIBentrySTDinterwordspacing

\bibitem{RFB15a}
\BIBentryALTinterwordspacing
O.~Ronneberger, P.Fischer, and T.~Brox, ``U-net: Convolutional networks for
  biomedical image segmentation,'' in \emph{Medical Image Computing and
  Computer-Assisted Intervention (MICCAI)}, ser. LNCS, vol. 9351.\hskip 1em
  plus 0.5em minus 0.4em\relax Springer, 2015, pp. 234--241, (available on
  arXiv:1505.04597 [cs.CV]). [Online]. Available:
  \url{http://lmb.informatik.uni-freiburg.de/Publications/2015/RFB15a}
\BIBentrySTDinterwordspacing

\bibitem{kingma2017adam}
D.~P. Kingma and J.~Ba, ``Adam: A method for stochastic optimization,'' 2017.

\bibitem{10.1109/JSTARS.2021.3074068}
I.~de~Gélis, A.~Colin, and N.~Longépé, ``Prediction of categorized sea ice
  concentration from sentinel-1 sar images based on a fully convolutional
  network,'' \emph{IEEE Journal of Selected Topics in Applied Earth
  Observations and Remote Sensing}, vol.~14, pp. 5831--5841, 2021.

\bibitem{1701.04128}
W.~Luo, Y.~Li, R.~Urtasun, and R.~Zemel, ``Understanding the effective
  receptive field in deep convolutional neural networks,'' 2017.

\bibitem{DBLP:journals/corr/abs-1206-5533}
\BIBentryALTinterwordspacing
Y.~Bengio, ``Practical recommendations for gradient-based training of deep
  architectures,'' \emph{CoRR}, vol. abs/1206.5533, 2012. [Online]. Available:
  \url{http://arxiv.org/abs/1206.5533}
\BIBentrySTDinterwordspacing

\end{thebibliography}


\begin{IEEEbiography}[{\includegraphics[width=1in,height=1in,clip,keepaspectratio]{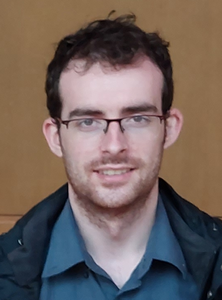}}]{Aurélien COLIN}
After graduating from the IMT Atlantique as an engineer, he began a Ph.D and is currently studying the segmentation of various meteorological and ocean phenomena on Synthetic Aperture Radar using deep learning models.
\end{IEEEbiography}

\vskip -2\baselineskip plus -1fil

\begin{IEEEbiography}[{\includegraphics[width=1in,height=1in,clip,keepaspectratio]{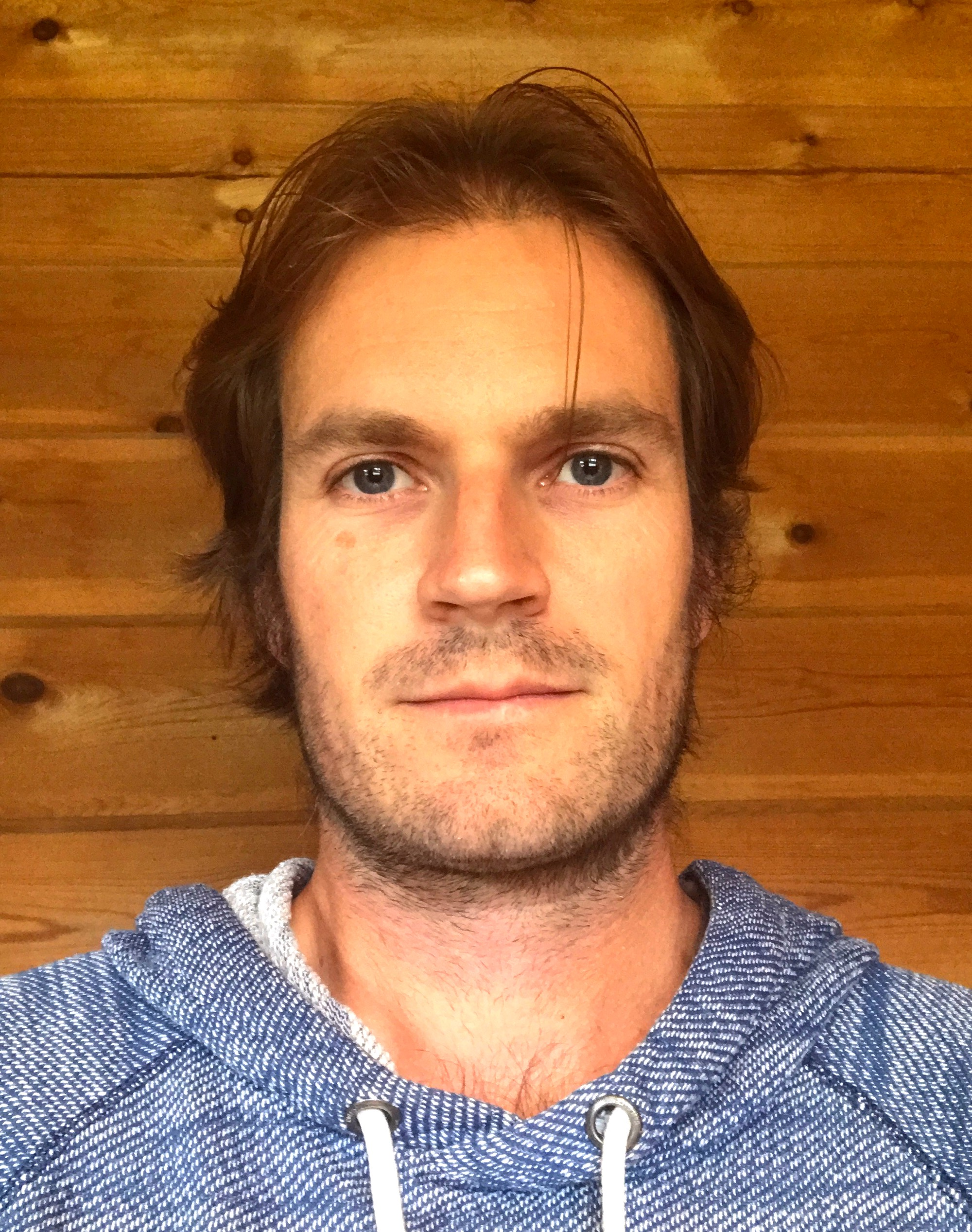}}]{Pierre TANDEO}
was born in France in 1983. He received the M.S. degree in applied statistics from Agrocampus Ouest, Rennes, France, and the Ph.D. degree from the Oceanography from Space Laboratory at IFREMER, Brest, France, in 2010. Then, he spent two years as a Postdoctoral Researcher with the Atmospheric Science Research Group, University of Corrientes, Argentina, and three years at Télécom Bretagne, Brest, France. Since 2015, he is an associate professor at IMT Atlantique, Brest, France, and a researcher at Lab-STICC, CNRS, France. Since 2019, he is an associate researcher at the Data Assimilation Research Team, RIKEN Center for Computational Science, Kobe, Japan. His main research interests are focused on IA, data assimilation, and inverse problems for geophysics.
\end{IEEEbiography}

\vskip -2\baselineskip plus -1fil

\begin{IEEEbiography}[{\includegraphics[width=1in,height=1in,clip,keepaspectratio]{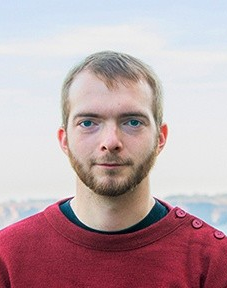}}]{Charles PEUREUX}
is a research engineer at CLS since 2020 where he is mainly involved with SAR oceanography. He previously worked at IFREMER where he defended his thesis on the observation and modelling of the directional properties o short gravity waves. Thereafter, he worked on the SKIM satellite mission, especially for the production of high resolution sea state numeral simulations.
\end{IEEEbiography}

\vskip -2\baselineskip plus -1fil

\begin{IEEEbiography}[{\includegraphics[width=1in,height=1in,clip,keepaspectratio]{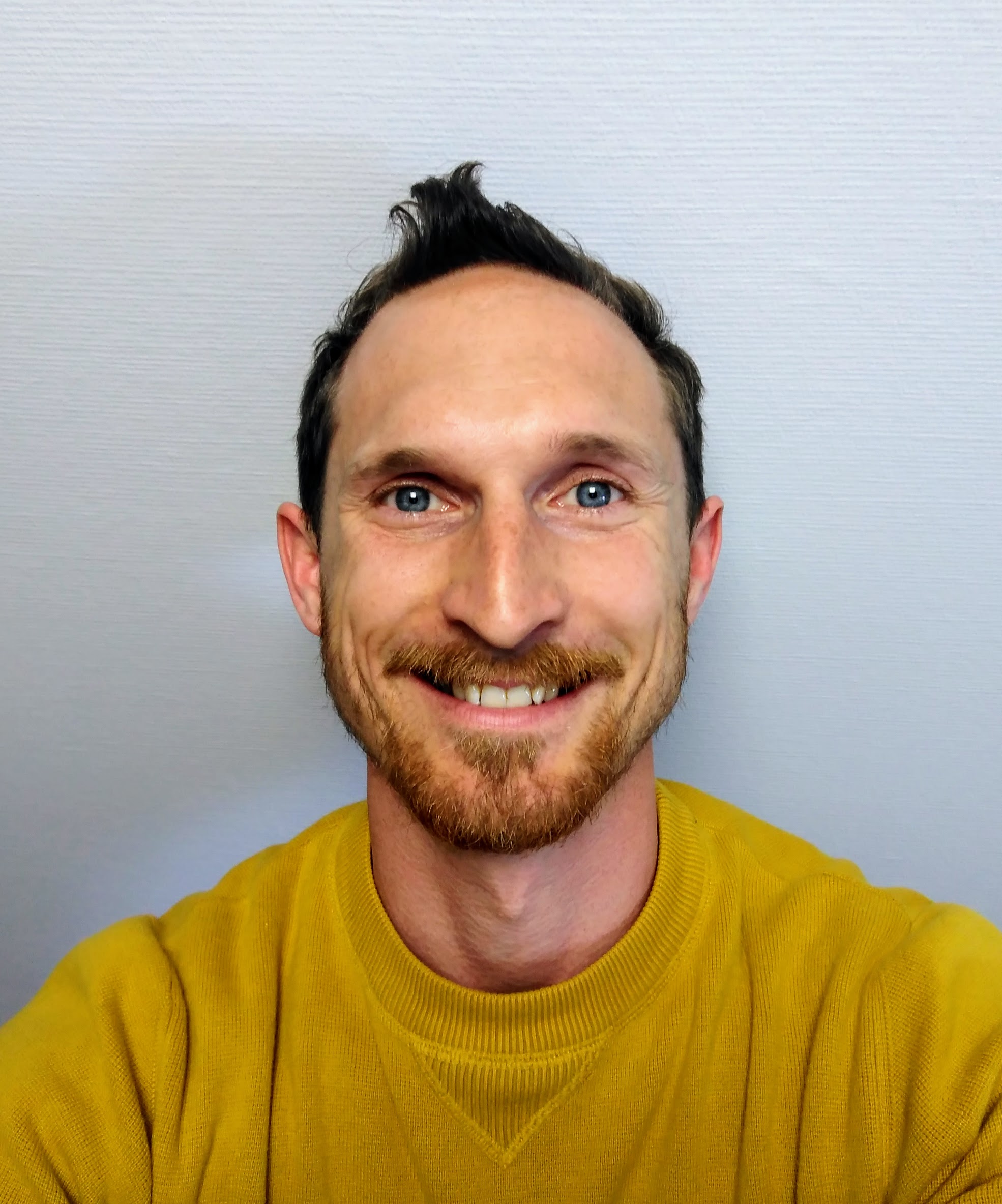}}]{Romain HUSSON}
is a scientist and project engineer at CLS, in the Environmental Monitoring and Climate Business Unit. After working as intern at NASA JPL and as YGT at ESRIN on SAR wind-wave-current activities, he completed in 2012 his PhD in Physical Oceanography at CLS Brest developing methods to estimate synthetic swell field from SAR wave mode observations. Since then, he has been involved in several applicative and R\&D projects such as wind/wave products for marine renewable energy and contributed to several ESA and European projects such as Cal/Val activities for ENVISAT and S-1 mission performance center as Level-2 expert. 
\end{IEEEbiography}

\vskip -2\baselineskip plus -1fil

\begin{IEEEbiography}[{\includegraphics[width=1in,height=1in,clip,keepaspectratio]{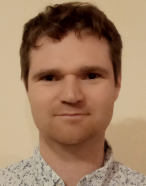}}]{Nicolas LONGEPE} received the M.Eng. degree in electronics and communication systems and the M.S. degree in electronics from the National Institute for the Applied Sciences, Rennes, France, in 2005, and the Ph.D. degree from the University of Rennes I, Rennes, France, in 2008. From 2007 to 2010, he was with the Earth Observation Research Center, Japan Aerospace Exploration Agency (JAXA), Tsukuba, Japan. From 2010 to 2020, he was with Collecte Localization Satellites (CLS), Plouzané, France, where he was a Research Engineer with the Space Observation Division. Since September 2020, he has been an Earth Observation Data Scientist at the $\Phi$-Lab Explore Office, European Space Research Institute (ESRIN), European Space Agency (ESA), Frascati, Italy. He has been working on the development of innovative SAR-based applications for environmental and natural resource management (ocean, mangrove, land and forest cover, soil moisture, snow cover, and permafrost) and for maritime security (oil spill, sea ice, iceberg, and ship detection/tracking). His main research interests include (EO) SAR remote sensing and digital technologies, such as machine (deep) learning.
\end{IEEEbiography}

\vskip -2\baselineskip plus -1fil

\begin{IEEEbiography}[{\includegraphics[width=1in,height=1in,clip,keepaspectratio]{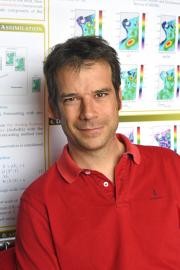}}]{Ronan FABLET}
is a Professor at IMT Atlantique and a research scientist at Lab-STICC in the field of Data Science and Computational Imaging. He is quite involved in interdisciplinary research at the interface between data science and ocean science, especially space oceanography and marine ecology. His current research interests include deep learning for dynamical systems and applications to the understanding, analysis, simulation and reconstruction of ocean dynamics, especially using satellite ocean remote sensing data.
\end{IEEEbiography}




\end{document}